\documentclass{article}
\usepackage[T1]{fontenc}
\usepackage{kernelmosaic-preprint}
\usepackage{amsmath,amssymb,booktabs,graphicx,array}
\usepackage{placeins}
\usepackage[authoryear,round]{natbib}
\setcitestyle{citesep={;},aysep={,},yysep={;}}
\renewcommand{\bibfont}{\small}
\usepackage[hidelinks]{hyperref}
\usepackage{xurl}

\newcommand{\kmdata}{KernelMosaic}
\newcommand{\kmframework}{KernelMosaic Framework}
\title{KernelMosaic: An Open Dataset and Agentic Framework for CUDA Kernel Generation}
\author{\small
  \href{https://openreview.net/profile?id=\string~Ningyuan_Xi1}{Ningyuan Xi}\textsuperscript{1,2,*}\quad
  \href{https://openreview.net/profile?id=\string~Yuhang_Lai1}{Yuhang Lai}\textsuperscript{1,2,*}\quad
  \href{https://openreview.net/profile?id=\string~Ning_Miao3}{Ning Miao}\textsuperscript{1,2}
}
\date{}
\hypersetup{%
  pdftitle={KernelMosaic: An Open Dataset and Agentic Framework for CUDA Kernel Generation},
  pdfauthor={Ningyuan Xi, Yuhang Lai, Ning Miao}
}
\begin{document}
\maketitle
\enlargethispage{10pt}
\begingroup
\makeatletter
\long\def\@makefntext#1{\parindent=0pt\noindent\@makefnmark\,#1}
\makeatother
\footnotetext[1]{Department of Data Science, City University of Hong Kong;~
  \textsuperscript{2}\,Hong Kong Institute of AI for Science, City University of Hong Kong.
  Correspondence to: Ning Miao (\href{mailto:ningmiao@cityu.edu.hk}{ningmiao@cityu.edu.hk}).
  \par\textsuperscript{*}Equal contribution; author order determined by a coin toss.}
\endgroup
\begin{abstract}
Training large language models to generate efficient CUDA kernels requires supervision spanning diverse computational requirements and iterative code optimization. Constructing this supervision is challenging because of heterogeneous specifications, and useful implementations emerge through iterative complicated testing and refinement. We introduce KernelMosaic, an open dataset pairing diverse CUDA tasks with verified solutions and trajectories. Human-guided review and diversity-aware selection curate tasks expressed as natural-language descriptions, reference programs, mathematical specifications, and their intersections. To build the solutions and trajectories, our agentic framework guides a strong LLM through implementation, correctness repair, and performance optimization with reusable CUDA guidance and execution feedback. Supervised fine-tuning of a small open-source model on KernelMosaic raises correctness from 4\%, 0\%, and 0\% to 88\%, 82\%, and 16\% on KernelBench L1, L2, and L3, respectively, while increasing the fraction of tasks with kernels faster than PyTorch eager at every level. Further training experiments show that supervision from the curated trajectories substantially outperforms direct responses and promotes sustained testing and revision. These results demonstrate that KernelMosaic enables a general-purpose model to achieve substantial gains in CUDA kernel generation.
\end{abstract}
\section{Introduction}
Efficient GPU kernels connect mathematical computation to the hardware that executes it. Developing them requires reconciling numerical requirements with parallel execution, memory access, and software interfaces. Automating this work with large language models requires reusable supervision covering diverse computational tasks and the iterative decisions that produce correct and efficient implementations.

Existing resources demonstrate several approaches to kernel-generation supervision: KernelBook provides paired implementations~\citep{paliskara2025kernelbook}, ConCuR emphasizes concise reasoning~\citep{kong2025concur}, and CUDA Agent and Nemotron-SFT-CUDA-v1 use agentic execution to develop solutions~\citep{dai2026cudaagent,nvidia2026nemotroncuda}. Constructing supervision across independently sourced natural-language, programmatic, and mathematical tasks introduces a joint curation and generation problem. Specifications differ in completeness and computational content, while their solutions require different evaluation procedures. A reusable collection must therefore preserve computational intent, maintain task diversity, and support the generation of tested implementations and informative trajectories.

Two challenges follow. First, task curation must resolve missing context and reduce redundancy without discarding distinct computations or changing their meaning. Second, solution generation must produce both reliable implementations and useful supervision of the iterative process. Correctness repair and performance optimization involve successive attempts; recording these attempts together with their feedback captures how implementation choices evolve into a working solution.

\begin{figure}[t]
\centering
\includegraphics[width=\linewidth]{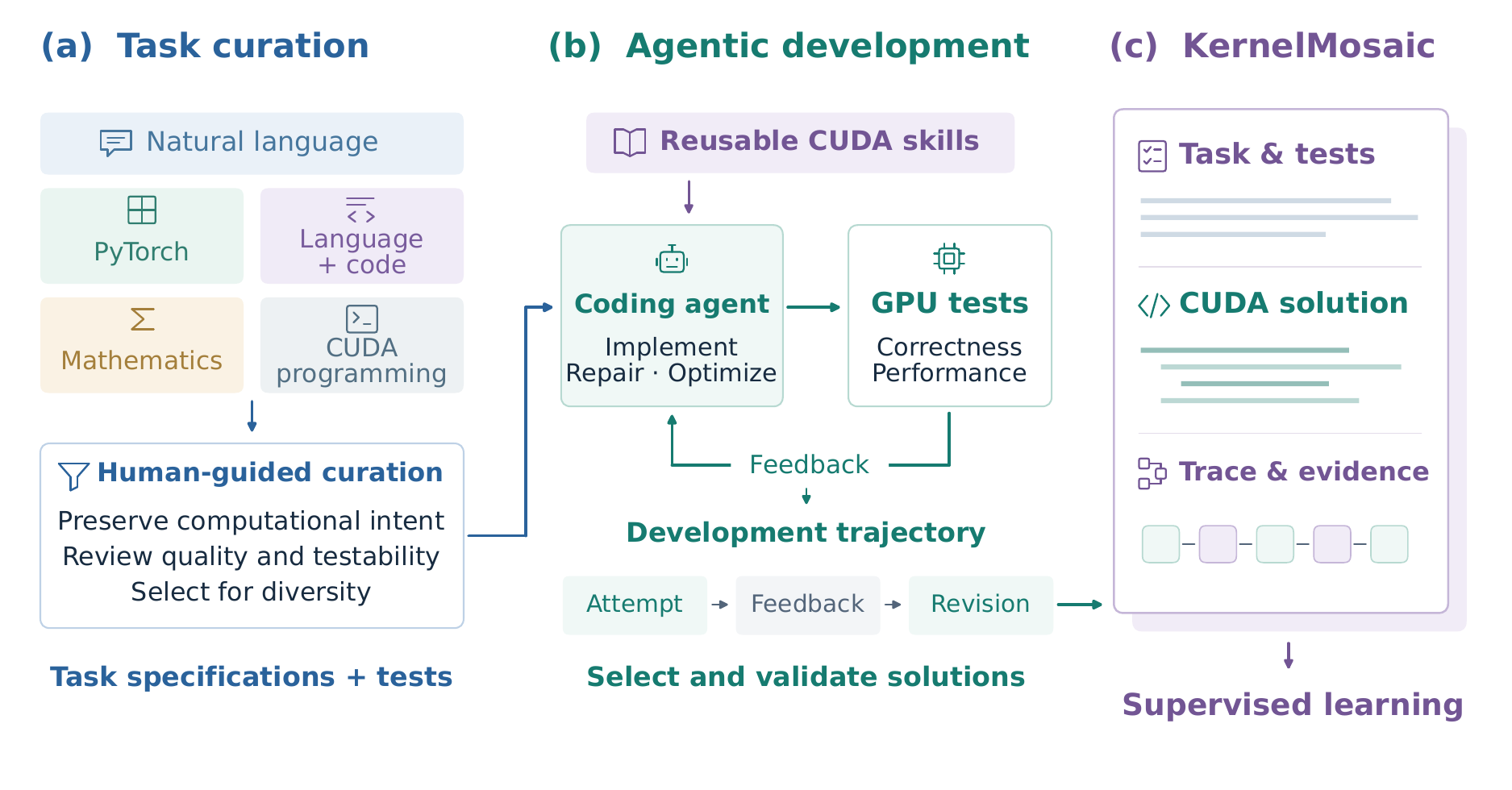}
\caption{Construction of \kmdata{}. \textbf{(a)} Human-guided curation preserves computational intent, reviews quality and testability, and selects for diversity across specification forms. \textbf{(b)} The agentic framework equips a coding agent with reusable CUDA skills and correctness and performance feedback to develop implementations and record attempts and observations. \textbf{(c)} The dataset links tasks and tests with selected CUDA solutions, trajectories, and evaluation evidence for supervised learning.}
\label{fig:pipeline}
\end{figure}

To address these challenges, we introduce \kmdata{}, an open dataset of 14,566 CUDA tasks from 41 sources, paired with verified solutions and trajectories (Figure~\ref{fig:pipeline}). Human-guided review improves specification quality, while diversity-aware selection broadens computational coverage. The \kmframework{} then constructs the solutions and trajectories by guiding a strong LLM through implementation, correctness repair, and performance optimization. Reusable CUDA guidance and execution feedback support this iterative development, and the recorded interactions provide supervision for training kernel-generation models.

We demonstrate the dataset's utility by fine-tuning Qwen3.5-9B~\citep{qwen35}. On KernelBench~\citep{ouyang2025kernelbench} L1, L2, and L3, correctness increases from 4\%, 0\%, and 0\% to 88\%, 82\%, and 16\%, respectively, with more tasks yielding kernels faster than PyTorch eager at every level. In a separate comparison using 1,000 examples and one training epoch, trajectory supervision yields 31 correct solutions, compared with one for direct-response supervision. Trajectory-trained models also engage in more sustained testing and revision, connecting the construction of training data with improved kernel-development behavior.

Our contributions are threefold. \textbf{First,} we introduce \kmdata{}, an open dataset combining diverse task specifications with verified CUDA solutions, trajectories, tests, and evaluation evidence. \textbf{Second,} we develop an agentic framework for constructing and extending this dataset through correctness repair and performance optimization. \textbf{Third,} we establish the dataset's utility for supervised learning through substantial benchmark gains and analyses of how trajectory supervision improves kernel-generation behavior.

\section{Related work}
\paragraph{Kernel datasets and learning.}
GPU programming datasets span several supervision formats (Table~\ref{tab:dataset-comparison}). LS-CAT pairs CUDA kernels with autotuning measurements~\citep{bjertnes2021lscat}, while CASS aligns source and assembly across NVIDIA and AMD platforms~\citep{heakl2026cass}. KernelBook supplies paired PyTorch and Triton programs~\citep{paliskara2025kernelbook}; AutoTriton and ConCuR construct code-and-reasoning supervision for Triton and CUDA, respectively~\citep{li2025autotriton,kong2025concur}. CUDA Agent combines synthesized tasks with an agentic reinforcement-learning environment~\citep{dai2026cudaagent}. Kevin learns through multi-turn reinforcement learning~\citep{baronio2026kevin}, CUDA-L1 through contrastive optimization~\citep{li2026cudal1}, and SparseRL targets sparse-matrix kernels~\citep{wang2026sparserl}. \kmdata{} brings heterogeneous specifications together with solutions and revision trajectories for supervised learning.

\paragraph{Agentic kernel generation.}
CUDA-LLM uses feature search and reinforcement to refine CUDA programs~\citep{chen2025cudallm}. CudaForge combines coding and judging agents with hardware feedback~\citep{zhang2025cudaforge}, while CUDA-Harness couples natural-language specifications with synthesized verification and iterative refinement~\citep{fan2026cudaharness}. Atrex-Kernel-Agent uses a GPU knowledge base~\citep{yang2026atrex}; ReGraphT transfers optimization experience to smaller models through retrieval and search over reasoning graphs~\citep{gong2026regrapht}. Dr.\ Kernel introduces KernelGYM for distributed evaluation, multi-turn trajectory collection, and reinforcement learning~\citep{liu2026drkernel}. AutoGuide structures execution feedback for parallel-program mapping optimization~\citep{wei2025agentinterfaces}. Our framework organizes tools, execution feedback, and domain guidance around dataset construction across multiple task representations.

\paragraph{Evaluation and task coverage.}
KernelBench measures functional correctness and speed across PyTorch workloads of increasing complexity~\citep{ouyang2025kernelbench}. TritonBench evaluates Triton generation on real-world operators and tasks aligned with PyTorch interfaces~\citep{li2025tritonbench}. Atrex-Bench derives workloads from production traces~\citep{yang2026atrex}. We curate training coverage through source-aware review and semantic selection using Qwen3 embeddings~\citep{zhang2025qwenembedding}.

\section{Task collection and curation}
\label{sec:tasks}
Task curation determines which computational requirements the dataset represents and whether those requirements can support meaningful implementation and evaluation. Our method combines source-aware construction, human-guided quality assessment, diversity selection, and preparation of executable tests.

\subsection{Preserving heterogeneous specifications}
We collect reference programs, existing instructions, paired implementations, and mathematical specifications. Each task describes a target computation, the requested CUDA implementation, and its source. Five categories capture the principal input forms and programming settings: natural language, PyTorch code, mixed language and code, algebraic or matrix specifications, and CUDA programming tasks such as source completion.

Construction follows the information supplied by each source. Translation tasks expose the reference program and withhold the target implementation. Completion tasks preserve the surrounding code needed to infer missing behavior. Mathematical tasks specify dimensions, domains, numerical types, and outputs while leaving implementation choices open. Across these forms, interface conventions and boundary conditions remain part of the specification. This treatment preserves meaningful differences between tasks while giving each a self-contained implementation objective.

\subsection{Human-guided quality assessment}
Human inspection and AI-assisted review identify ambiguous requirements, missing context, unsuitable computational targets, and unnecessary solution guidance. Review findings refine source-specific construction rules, allowing a correction to improve other examples with the same underlying issue. Known semantic defects and non-CUDA targets are excluded. For tasks not reviewed individually, admission relies on source-level audits or upstream validation with verified input extraction. Appendix~\ref{app:construction-details} specifies these admission rules.

Review also distinguishes substantive variation in specifications from differences in presentation. Mixed language-and-code tasks combine program context with requirements expressed in prose. Review corrects category labels, repairs incomplete specifications, and excludes unsuitable tasks while retaining differences in computational requirements and input form.

\subsection{Diversity-aware selection}
We augment underrepresented mathematical inputs with source-grounded and synthetic specifications, and remove normalized duplicates. We then use Qwen3-Embedding-8B~\citep{zhang2025qwenembedding} for weighted farthest-first selection over the combined pool. Each task receives a weight $w_i$ combining source provenance, recorded review scores, input category, mathematical reasoning, and CUDA relevance.

For normalized task embeddings $z_i$ and retained set $S$, the similarity of a task to its closest retained representative is
\begin{equation}
\operatorname{coverage}(i,S)=\max_{j\in S}z_i^\top z_j.
\end{equation}
Starting with the highest-weight task, selection repeatedly adds the uncovered task maximizing $w_i[1-\operatorname{coverage}(i,S)]$. A task is covered when its similarity to a retained representative reaches the cutoff. This direct comparison prevents a chain of moderately similar tasks from obscuring an uncovered example. Weights, thresholds, and the additional mathematical-task selection are given in Appendix~\ref{app:construction-details}.

\subsection{Preparing executable evaluation}
We recover upstream references and tests, and supplement them with generated test cases and implementation interfaces. Evaluation resources specify dependencies, expected outputs, and numerical tolerances. Task clarity and evaluability are assessed separately because a well-defined computation may still require additional inputs or fixtures to execute.

We assess test suites through reference execution and candidate controls, using valid implementations and deliberately corrupted outputs to probe checker behavior. Evaluation records distinguish candidate tests, reference-only checks, and restricted-buffer comparisons. Compilation, reference execution, and candidate correctness are recorded separately. Tasks without a usable evaluation procedure are excluded from trajectory generation. These resources provide feedback for implementation development.

\section{Agentic kernel and trajectory generation}
\label{sec:framework}
The framework builds the dataset's solutions and development trajectories. A model-facing harness, task-specific evaluation, and a revision controller jointly develop and select implementations across specification forms and model configurations.

\subsection{Reusable CUDA guidance}
The harness builds on OpenCode~\citep{opencode}, providing tools for code editing, execution, and inspection. Reusable CUDA skills organize guidance on numerical behavior, memory access, parallel computation, launch overhead, fusion, and performance diagnosis. The harness exposes this guidance through callable skills that connect optimization knowledge to task-specific implementation decisions.

Skills support direct invocation and optional routing. Routing associates guidance with development stages and observed failures, such as numerical mismatches or repeated compilation errors, and suggests relevant skills for the next revision.

\subsection{Implementation, repair, and optimization}
Given task $x_i$ and development history $h_t$, the agent proposes a candidate implementation $c_t$. \mbox{Evaluation} returns compilation outcomes, numerical checks, and performance measurements. This feedback guides repair until a candidate passes the correctness tests, after which further revisions seek improved performance. Previously valid implementations remain available throughout optimization, and resource constraints bound the development process.

Let $\mathcal{C}_i$ denote the candidate set, $v_i^{\mathrm{pub}}(c)$ indicate correctness on tests visible during development, and $s_i^{\mathrm{pub}}(c)$ denote the corresponding performance score. When at least one candidate passes, the controller selects
\begin{equation}
  c_i^\star \in
  \operatorname*{arg\,max}_{c \in \mathcal{C}_i:\,v_i^{\mathrm{pub}}(c)=1}
  s_i^{\mathrm{pub}}(c).
\end{equation}
Tasks without a passing candidate retain a failure outcome. Held-out tests evaluate the selected implementation only after development; their inputs are never exposed to the agent.

\subsection{Qualification and trajectory construction}
Solutions are admitted by the correctness checks specified in each task's evaluation policy. The policy determines the use of public and held-out tests and defines requirements on output structure, input mutation, and numerical tolerances. Performance is measured against a specified reference on a defined workload and device. Correctness and speed are recorded separately.

The framework records model responses, candidate programs, revisions, and evaluation feedback alongside the selected implementation. The resulting trajectory captures successful changes as well as failed attempts that informed later decisions. Task identity and evaluation outcomes connect this history to the computation it addresses. These associations support learning from both final solutions and the process of developing them.

The dataset supports solution-based and trajectory-based training views. Our supervised study uses trajectories segmented into complete assistant decisions and associated tool observations, as detailed in Appendix~\ref{app:training-data}. The same framework supports additional data generation and kernel development with other models. Implementation and optional explanation-generation details appear in Appendix~\ref{app:construction-details}.

\section{The KernelMosaic dataset}
\label{sec:dataset}
\subsection{Tasks, solutions, and development trajectories}
\kmdata{} associates each task with its specification, source, available tests, generated implementations, and trajectory. Evaluation outcomes identify the selected solution and its assessment conditions. Tasks support fresh generation experiments, solutions provide implementation supervision, and trajectories expose repair and optimization behavior. Failed intermediate attempts preserve the feedback that motivated subsequent revisions.

\subsection{Dataset composition}
The dataset contains 14,566 tasks from 41 sources, each paired with a successful trajectory generated by Qwen3.8-Flash~\citep{qwen38flash} under fixed token and time budgets and an implementation that passes the task's evaluation checks. Table~\ref{tab:dataset-comparison} compares resources for CUDA, Triton, and cross-platform GPU programming. Appendix~\ref{app:dataset-accounting} gives category counts and generation budgets.

\begin{table}[!htbp]
\caption{Related datasets for GPU kernel learning. Scale uses each resource's native unit. Each of the 14,566 tasks in \kmdata{} has an associated development trajectory. Mixed denotes multiple source and assembly languages.}
\label{tab:dataset-comparison}
\centering\small
\setlength{\tabcolsep}{4pt}
\renewcommand{\arraystretch}{1.2}
\begin{tabular}{@{}>{\raggedright\arraybackslash}p{.405\linewidth} l r l@{}}
\toprule
Dataset & Domain & Scale & Data / supervision \\
\midrule
LS-CAT \citep{bjertnes2021lscat} & CUDA & 19,683 kernels & Code and runtimes \\
KernelBook \citep{paliskara2025kernelbook} & Triton & 18,162 pairs & PyTorch--Triton pairs \\
AutoTriton SFT \citep{li2025autotriton} & Triton & 14,102 examples & Code and reasoning \\
ConCuR \citep{kong2025concur} & CUDA & 4,892 examples & Code and reasoning \\
CASS \citep{heakl2026cass} & Mixed & 60,694 samples & Cross-platform pairs \\
CUDA-Agent-Ops-6K \citep{dai2026cudaagent} & CUDA & 6,000 tasks & PyTorch tasks, tests \\
\midrule
\kmdata{} & CUDA & 14,566 tasks & Solutions, trajectories \\
\bottomrule
\end{tabular}
\end{table}

The dataset comprises 13,815 training and 751 validation task--trajectory records across all five input categories. PyTorch programs account for 68.4\% of tasks, alongside CUDA programming, mixed language-and-code, algebraic, and natural-language specifications. Appendix~\ref{app:dataset-analysis} examines computational vocabulary, category coverage, source profiles, and supervision.

\subsection{Learning from development trajectories}
\begin{figure}[t]
\centering
\includegraphics[width=\linewidth]{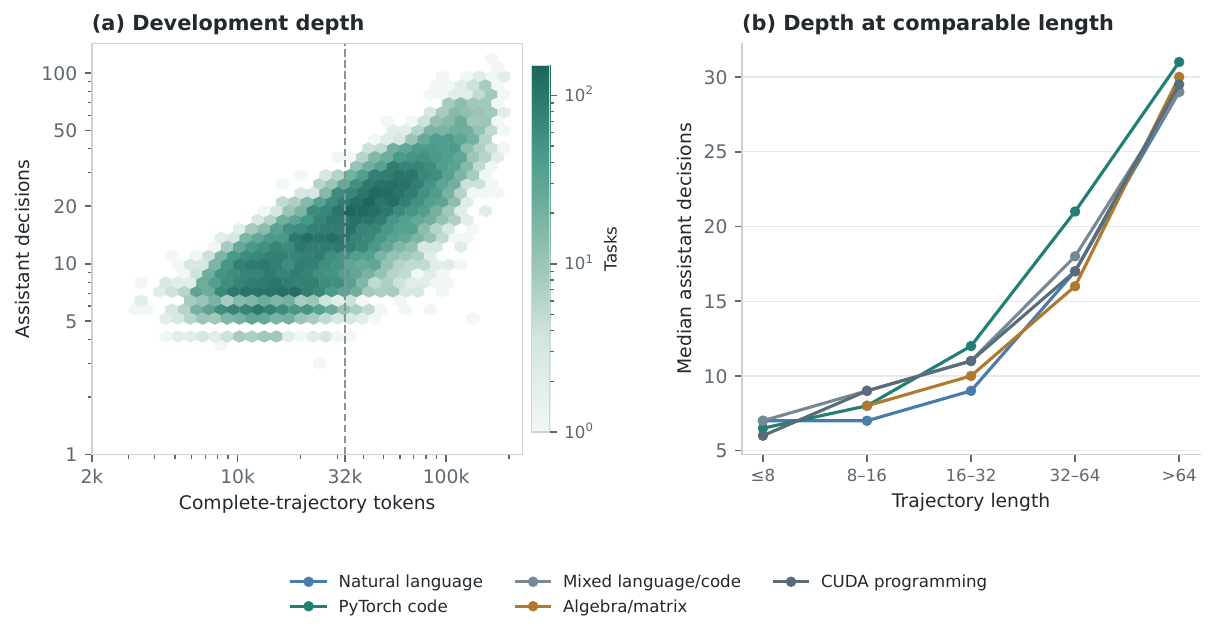}
\caption{Length and interaction depth of the 14,566 trajectories. \textbf{(a)} Density of complete-trajectory tokens and assistant decisions on logarithmic scales; the dashed line marks the 32,768-token training context. \textbf{(b)} Median decision counts by input category within token-length intervals ($\times$1,024). Intervals are right-closed; points with fewer than 20 tasks are omitted. Length includes the complete context and historical reasoning.}
\label{fig:development-depth}
\end{figure}

Trajectories contain a median of 14 assistant decisions, and 47.4\% exceed the 32,768-token training context limit (Figure~\ref{fig:development-depth}a). Interaction depth also varies at comparable token lengths: within 32,768--65,536 tokens, median decision counts range from 16 for algebraic tasks to 21 for PyTorch tasks; above 65,536 tokens, all five input categories have medians between 29 and 31 (Figure~\ref{fig:development-depth}b). These trajectories provide supervision across multiple rounds of implementation and revision.

We divide long trajectories at complete assistant-decision and tool-interaction boundaries, producing 37,242 training and 2,010 validation windows. Each window preserves the task and relevant history; only newly assigned assistant decisions contribute to the loss. Repeated history and tool observations provide context without duplicating supervision. Appendix~\ref{app:protocol} specifies window construction, and Appendix~\ref{app:examples} illustrates tasks and a measured LayerNorm revision.

\section{Experiments}
\label{sec:experiments}
We evaluate the utility of \kmdata{} for learning CUDA kernel generation. KernelBench covers individual operations, operation sequences, and model-level computations; we report correctness and performance at each level~\citep{ouyang2025kernelbench}.

\subsection{Experimental setup}
We compare Qwen3.5-9B~\citep{qwen35} with full-parameter SFT on \kmdata{} for three epochs over 37,242 training windows. Two one-epoch variants with 1,000 examples each compare trajectory and direct-response supervision. All use a 32,768-token context limit and AdamW with a constant learning rate of $10^{-5}$. Appendix~\ref{app:protocol} provides the training configurations and supervision objective.

All four checkpoints use a common agentic evaluation protocol on KernelBench v0.1's 100 L1, 100 L2, and 50 L3 tasks. Each task requests CUDA C++ code through the reference module's Python interface and permits one trajectory with up to 262,144 generated tokens and 3,600 active seconds, excluding inference and infrastructure waits. Inference runs on B200 GPUs; candidate evaluation uses H200 GPUs.

A task is solved when an accepted candidate passes the numerical checks; its performance is the largest retained speedup among these candidates. Timing compares synchronized forward-call wall time against PyTorch eager, including Python dispatch. Appendix~\ref{app:kernelbench-initial} specifies correctness checks, timing, and budget accounting.

\paragraph{Metrics.}
For $N$ tasks, let $v_i$ indicate whether a task is solved and $s_i$ denote its best retained reference-to-candidate speedup. Following KernelBench, we report correctness and
\begin{equation}
\operatorname{fast}_{p}=\frac{1}{N}\sum_{i=1}^{N}
\mathbf{1}\{v_i=1\ \land\ s_i>p\},
\end{equation}
for $p\in\{1,2\}$. Each metric includes all tasks in the denominator, including unsuccessful and timed-out attempts. A correct candidate without a reliable timing measurement contributes to correctness but not to fast$_p$.

\begin{table}[t]
\caption{KernelBench correctness (C) and faster-than-PyTorch-eager coverage (fast$_1$), in percent. The 1k variants each use 1,000 training examples and one epoch. Published results use different evaluation protocols. Rows from \citet{li2025autotriton} generate Triton; the others generate CUDA. -- denotes an unreported result. Appendix~\ref{app:published-kernelbench} gives protocol details and additional speedup thresholds.}
\label{tab:main}
\centering\small
\setlength{\tabcolsep}{5pt}
\renewcommand{\arraystretch}{1.07}
\begin{tabular}{@{}l l rr rr rr@{}}
\toprule
Method / model & Inference & \multicolumn{2}{c}{L1} & \multicolumn{2}{c}{L2} & \multicolumn{2}{c}{L3} \\
\cmidrule(lr){3-4}\cmidrule(lr){5-6}\cmidrule(l){7-8}
 & & C & fast$_1$ & C & fast$_1$ & C & fast$_1$ \\
\midrule
\multicolumn{8}{@{}l}{\textit{Open-weight models}} \\
Qwen3-8B \citep{li2025autotriton} & Single-turn & 16 & 9 & 16 & 8 & 14 & 4 \\
DeepSeek-R1-0528 \citep{li2025autotriton} & Single-turn & 35 & 7 & 42 & 28 & 26 & 14 \\
\addlinespace[3pt]
\multicolumn{8}{@{}l}{\textit{Kernel-specialized open-weight models}} \\
AutoTriton-8B \citep{li2025autotriton} & Single-turn & 36 & 10 & 45 & 17 & 20 & 10 \\
Kevin-32B \citep{kong2025concur} & Single-turn & 50 & 16 & 46 & 27 & -- & -- \\
KernelCoder-32B \citep{kong2025concur} & Single-turn & 58 & 17 & 59 & 39 & -- & -- \\
\addlinespace[3pt]
\multicolumn{8}{@{}l}{\textit{Proprietary models and systems}} \\
GPT-4o \citep{li2025autotriton} & Single-turn & 15 & 3 & 5 & 3 & 8 & 4 \\
Claude Sonnet 4 \citep{li2025autotriton} & Single-turn & 33 & 11 & 26 & 10 & 18 & 2 \\
Claude Opus 4.5 \citep{dai2026cudaagent} & Agentic & 96 & 88 & 98 & 97 & 88 & 82 \\
Gemini 3 Pro \citep{dai2026cudaagent} & Agentic & 95 & 90 & 93 & 91 & 80 & 76 \\
CudaForge (o3) \citep{zhang2025cudaforge} & Multi-turn & 96 & 54 & 100 & 89 & 96 & 68 \\
CUDA Agent (Seed1.6) \citep{dai2026cudaagent} & Agentic & 100 & 99 & 100 & 100 & 94 & 94 \\
\midrule
\multicolumn{8}{@{}l}{\textit{Our Qwen3.5-9B comparison}} \\
Qwen3.5-9B & Agentic & 4 & 0 & 0 & 0 & 0 & 0 \\
\quad + Single-turn SFT (1k) & Agentic & 1 & 1 & 0 & 0 & 0 & 0 \\
\quad + Trajectory SFT (1k) & Agentic & 23 & 13 & 8 & 4 & 0 & 0 \\
\quad + Trajectory SFT (full) & Agentic & 88 & 64 & 82 & 62 & 16 & 12 \\
\bottomrule
\end{tabular}
\end{table}

\subsection{Kernel-generation performance}
Table~\ref{tab:main} reports results by level alongside published models and systems. The comparison covers general and kernel-specialized open-weight models, frontier proprietary models, and iterative systems. Within our Qwen3.5-9B comparison, fine-tuning increases L1 correctness from 4\% to 88\% and fast$_1$ from 0\% to 64\%. On L2, the original model produces no correct candidates, whereas the fine-tuned model achieves 82\% correctness and 62\% fast$_1$. On L3, correctness increases from 0\% to 16\% and fast$_1$ from 0\% to 12\%.

\begin{figure}[t]
\centering
\includegraphics[width=\linewidth,trim=0 2mm 0 0,clip]{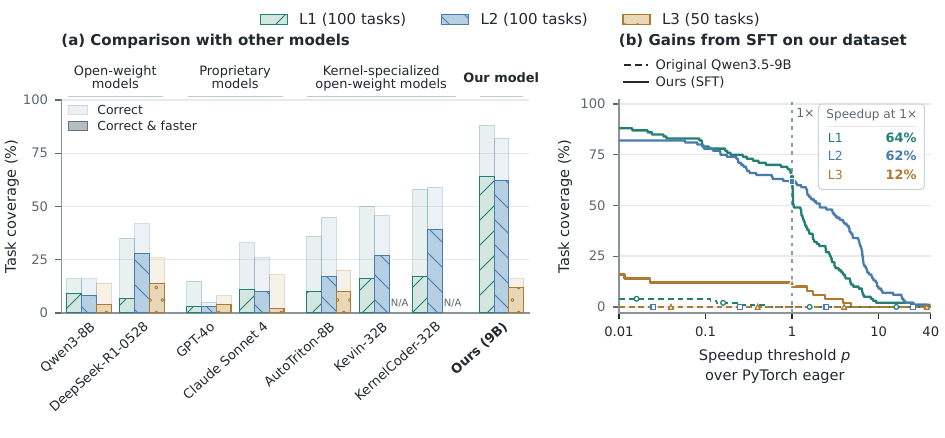}
\caption{KernelBench performance. \textbf{(a)} Correctness (pale background bars) and correct-and-faster coverage (hatched foreground bars; speedup $>1$) of selected models from Table~\ref{tab:main} and full-dataset SFT. Both layers start at zero; N/A denotes unreported results. Published baselines use direct generation or pass@1, while ours use agentic inference; budgets are not matched. \textbf{(b)} Original and SFT task coverage above speedup $p$. Colors identify L1/L2/L3 (100/100/50 tasks); our candidates are evaluated on H200.}
\label{fig:kernelbench-initial}
\end{figure}

Figure~\ref{fig:kernelbench-initial}a compares correctness and faster-than-eager coverage for selected models. We exclude the four frontier models because their performance has saturated on the benchmark. Figure~\ref{fig:kernelbench-initial}b shows performance across speedup thresholds. Fine-tuning yields fast$_2$ of 31\%, 51\%, and 6\% on L1, L2, and L3, respectively.

\subsection{Effect of supervision construction}
\label{sec:supervision-comparison}
\textbf{Trajectory SFT} uses 1,000 successful framework-generated trajectories containing reasoning, tool use, and revisions conditioned on execution feedback. \textbf{Direct SFT} uses 1,000 one-shot reasoning-and-code responses generated without the framework or correctness filtering. The cohorts are independently sampled; both models start from Qwen3.5-9B and train for one epoch with the same learning rate, global batch size, and context limit. Appendix~\ref{app:supervision-comparison} details cohort selection, token counts, and optimization.

\begin{figure}[t]
\centering
\includegraphics[width=\linewidth]{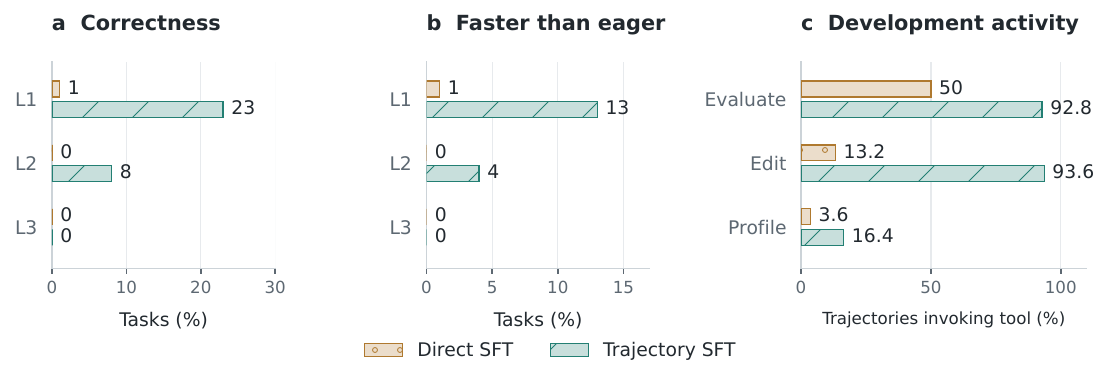}
\caption{Supervision from agentic trajectories and direct responses. Both variants train Qwen3.5-9B for one epoch on 1,000 independently sampled examples. \textbf{(a--b)} KernelBench correctness and faster-than-eager coverage by level. \textbf{(c)} Fraction of the 250 evaluation trajectories invoking each tool. Trajectory supervision improves L1/L2 performance and supports more frequent evaluation and editing.}
\label{fig:sft-comparison}
\end{figure}

Across all 250 tasks, Trajectory SFT yields 31 correct solutions and 17 faster-than-eager solutions, versus one of each for Direct SFT (Table~\ref{tab:main}; Figure~\ref{fig:sft-comparison}a--b). The gains span L1 and L2; neither variant solves an L3 task. Evaluated agentic iteration thus produces more effective training data than direct responses in this comparison.

Trajectory supervision also promotes sustained testing and revision (Figure~\ref{fig:sft-comparison}c): evaluation occurs on 92.8\% of tasks and editing on 93.6\%, compared with 50.0\% and 13.2\% for Direct SFT. Median tool calls increase from 4 to 105. Appendix~\ref{app:sft-development} analyzes this activity by level.

\subsection{Model behavior during kernel development}
\label{sec:model-behavior}
\begin{figure}[t]
\centering
\includegraphics[width=\linewidth]{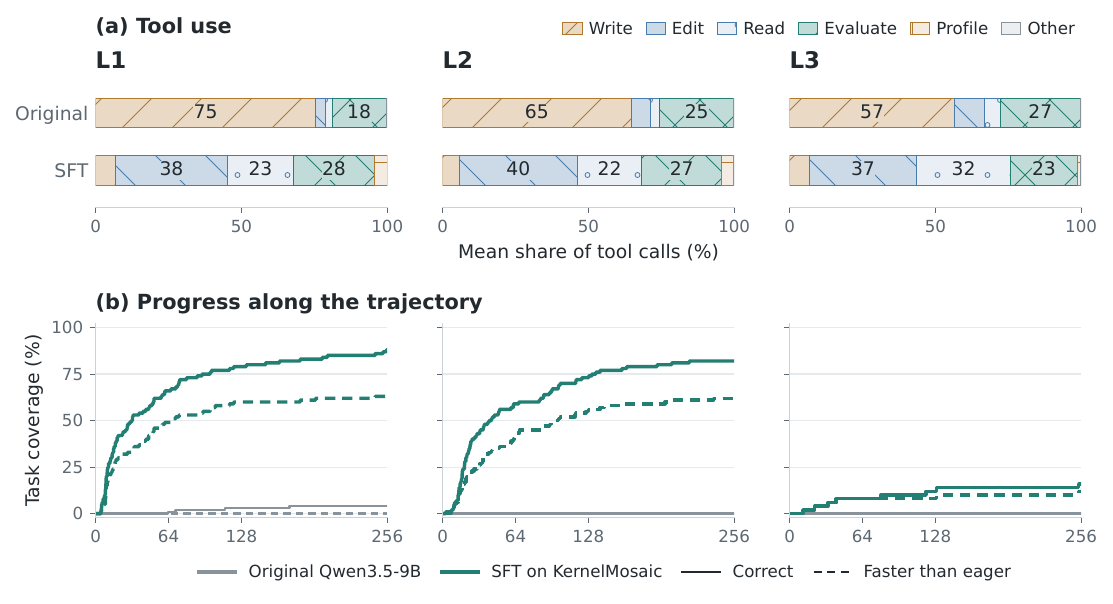}
\caption{Original and full-dataset SFT behavior on KernelBench. \textbf{(a)} Mean within-trajectory tool-use shares; Write creates or replaces files, while Edit modifies existing contents. \textbf{(b)} Correct (solid) and faster-than-eager (dashed) task coverage versus cumulative output tokens ($\times$1,024). Each panel includes all tasks at its level.}
\label{fig:kernelbench-behavior}
\end{figure}

Comparing the original checkpoint with full-dataset SFT across 500 benchmark trajectories, SFT shifts tool use from file writes toward editing, code inspection, and evaluation (Figure~\ref{fig:kernelbench-behavior}a). Repeated file contents account for 60.1\% of the original model's completed writes, compared with 2.6\% after SFT. Profiling appears in 87\%, 85\%, and 30\% of SFT trajectories on L1--L3, versus 6\%, 4\%, and 2\% for the original model.

By 65,536 output tokens, SFT solves 66\% of L1 and 59\% of L2 tasks, compared with 1\% and 0\% for the original model (Figure~\ref{fig:kernelbench-behavior}b). L3 correctness reaches 8\% at this point and 16\% at the full budget.

Development continues after correctness. SFT's first timed correct candidates outperform PyTorch eager on 49\% of L1 and 46\% of L2 tasks. Subsequent distinct implementations raise this coverage to 64\% and 62\%, respectively. On L3, six of the eight solved tasks exceed eager performance at their first timed correct candidate. Appendix~\ref{app:kernelbench-behavior} analyzes action sequences, recovery, resource use, and task-level optimization.

\FloatBarrier
\section{Conclusion}
\kmdata{} links diverse computational specifications with tested CUDA solutions and agentic trajectories. Its framework builds this supervision through iterative implementation, repair, and optimization. Fine-tuning Qwen3.5-9B raises KernelBench L1, L2, and L3 correctness from 4\%, 0\%, and 0\% to 88\%, 82\%, and 16\%, respectively, and faster-than-eager coverage from zero to 64\%, 62\%, and 12\%. With 1,000 examples and one epoch, training on framework-generated trajectories solves 31 benchmark tasks, compared with one for direct-response supervision. The learned behavior emphasizes evaluation and targeted editing, connecting the framework's data-construction process with effective iterative development. KernelMosaic enables a general-purpose model to achieve large gains in both correct and faster-than-eager CUDA generation through iterative agentic workflows.

\par
\label{maintextend}
\clearpage
\begingroup
\raggedright
\bibliographystyle{plainnat}
\bibliography{references}
\endgroup
\clearpage
\appendix
\section{Supervised training details}
\label{app:protocol}
We train Qwen3.5-9B by full-parameter supervised fine-tuning on development trajectories from \kmdata{}. The initialization is the checkpoint used in the original-model comparison.

\subsection{Trajectory representation and supervision}
\label{app:training-data}
Windows are constructed independently within the task-disjoint training and validation splits summarized in Table~\ref{tab:training-export}. Training covers all 41 sources and five input categories.

To accommodate long trajectories, we divide them into windows at complete decision boundaries. A decision comprises an assistant message and its associated tool observations, or an assistant message without a tool call. Each window retains the original task and all user constraints encountered so far. Historical assistant reasoning is removed while prior actions and tool observations remain as context. Complete target decisions are added within the 32,768-token limit. If the next target cannot fit, the oldest complete historical decisions are removed as needed. A target that still cannot fit with the required task context is excluded as a whole.

Each retained decision contributes supervision exactly once across windows. Newly assigned assistant reasoning, tool calls, and text are supervised; task context, tool observations, and repeated historical assistant content are masked. For window $z_i$ and binary target mask $m_{it}$, the objective is
\begin{equation}
\mathcal{L}(\theta)=-\frac{\sum_{i,t}m_{it}\log p_\theta(z_{it}\mid z_{i,<t})}{\sum_{i,t}m_{it}}.
\end{equation}
Each training example contains one complete exported window. Padding positions are excluded from the loss.

This procedure retains 252,371 of 253,044 decisions (99.7\%) and excludes 673 oversized decisions. Every source trajectory contributes at least one example. Of 39,252 windows, 10,796 (27.5\%) require removal of some historical context. Table~\ref{tab:training-export} reports the resulting trajectory, window, and token counts per copy of each split.

\begin{table}[ht]
\caption{Trajectory and token counts for supervised learning. Total tokens include context; supervised tokens contribute to the loss.}
\label{tab:training-export}
\centering\small
\begin{tabular}{lrrrr}
\toprule
Split & Trajectories & Windows & Total tokens & Supervised tokens \\
\midrule
Training & 13,815 & 37,242 & 1,002,791,260 & 416,424,437 \\
Validation & 751 & 2,010 & 54,137,600 & 22,460,320 \\
\midrule
Total & 14,566 & 39,252 & 1,056,928,860 & 438,884,757 \\
\bottomrule
\end{tabular}
\end{table}

\subsection{Optimization}
Table~\ref{tab:config} specifies the optimization settings. We evaluate the final checkpoint after three epochs; Appendix~\ref{app:kernelbench-initial} gives the inference and evaluation protocol.

\begin{table}[ht]
\caption{Full-dataset supervised fine-tuning configuration.}
\label{tab:config}
\centering\small
\begin{tabular}{p{0.42\linewidth}p{0.49\linewidth}}
\toprule
Parameter & Value \\
\midrule
Model / trainable parameters & Qwen3.5-9B / all 8,953,803,264 \\
Optimizer steps & 13,968 \\
Per-device batch / gradient accumulation & 1 / 1 \\
Global batch / training GPUs & 8 / eight B200 GPUs \\
Optimizer / learning rate / warm-up & AdamW / constant $10^{-5}$ / none \\
Gradient clipping / model and data seeds & 1.0 / 42 \\
Sequence length / packing & 32,768 / none \\
Precision / sharding / attention & BF16 / FSDP2 / FlashAttention-2 \\
Activation checkpointing & 26 of 32 layers \\
\bottomrule
\end{tabular}
\end{table}

\subsection{Trajectory and direct-response supervision}
\label{app:supervision-comparison}
The 1,000-example comparison evaluates two supervision-construction approaches, both using Qwen3.8-Flash to generate the training responses. For Trajectory SFT, we uniformly sample 1,000 trajectories from the 13,815-task training split and retain every associated decision window. The sample contains 2,519 windows, with 17,050 supervised decisions; 51 oversized decisions are excluded by the window-construction rule above. Direct SFT uses one-shot reasoning-and-code responses to curated task specifications, without tools, execution feedback, or correctness filtering. We preserve each complete system--user--assistant sequence and sample 1,000 of the 1,079 responses that fit within 32,768 tokens, from a source collection of 5,164 responses. Sampling uses seed 20260923 in both cases. The independently selected cohorts share 61 task IDs.

Both variants use the original Qwen3.5-9B initialization, full-parameter AdamW optimization, a constant learning rate of $10^{-5}$, global batch size eight, BF16, FSDP2, FlashAttention-2, and model/data seed 42. They train for one epoch with no packing or learning-rate warm-up. Trajectory targets follow the decision-level masking described above; Direct SFT supervises the complete assistant reasoning and answer while masking the system and user messages. Table~\ref{tab:supervision-training} reports the resulting training exposure. Trajectory SFT runs on four B200 GPUs with gradient accumulation two; Direct SFT runs on eight H200 GPUs with accumulation one.

\begin{table}[ht]
\caption{Training data and optimization for the one-epoch supervision comparison. Tokens are counted before repetition across epochs; total tokens include context.}
\label{tab:supervision-training}
\centering\small
\begin{tabular}{lrr}
\toprule
 & Direct SFT & Trajectory SFT \\
\midrule
Source examples & 1,000 & 1,000 \\
Sources & 20 & 23 \\
\addlinespace[3pt]
Natural-language tasks & 41 & 32 \\
PyTorch specifications & 31 & 677 \\
Mixed language and code & 264 & 117 \\
Algebraic or matrix specifications & 12 & 30 \\
CUDA programming tasks & 652 & 144 \\
\addlinespace[3pt]
Training windows & 1,000 & 2,519 \\
Total tokens & 12,594,890 & 66,957,025 \\
Supervised tokens & 11,416,743 & 28,947,222 \\
Optimizer steps & 125 & 315 \\
\bottomrule
\end{tabular}
\end{table}

\section{KernelBench evaluation protocol}
\label{app:kernelbench-initial}
We evaluate all four checkpoints on all 250 tasks in KernelBench v0.1: 100 L1, 100 L2, and 50 L3 tasks. Each model generates one trajectory per task.

\subsection{Task adaptation and inference}
The adapter preserves the upstream reference programs and input generators. It requests a CUDA C++ implementation through a Python module that maintains the reference constructor and forward interface. Tools support file reading, writing, editing, GPU evaluation, and profiling. Thinking and native context compaction are enabled with the default Qwen chat template. Each response permits 32,768 tokens, including a reasoning budget of 24,576 tokens. Task-level limits are 262,144 generated tokens and 3,600 active seconds, including a 300-second finalization reserve. The active clock excludes model inference and infrastructure waits; generated reasoning and response tokens count toward the token limit.

Checkpoints are served in pairs, each through four single-GPU replicas on B200 devices. The original and full-dataset SFT pair precedes the two 1,000-example variants. Compilation precedes GPU evaluation, which assigns each candidate exclusive access to one of eight H200 NVL devices.

\subsection{Correctness and timing}
A candidate must execute CUDA code and match the reference outputs on three input seeds (104729, 130363, 155921), with absolute and relative tolerances both 0.01. The evaluation tools return numerical and performance feedback during the trajectory.

Measurements use PyTorch 2.7.1 with CUDA 12.6. Before formal evaluation, we warm up the GPU until its operating frequency stabilizes around 1,785\,MHz. Forward-call timings synchronize the GPU and include Python dispatch. Constructors, input cloning, compilation, and numerical comparison are outside the timed interval. For each implementation, we additionally perform one warm-up pass and three measured passes over the three input seeds. The timing stage therefore executes twelve forward calls per implementation: three warm-up calls and nine measured calls. Numerical checks are repeated during measurement. Speedup is the median of the nine reference latencies divided by the median of the nine candidate latencies. Timing measurements are retained when the relative median absolute deviation is at most 25\% for both the reference and candidate.

\subsection{Trajectory scoring}
Scoring pools the candidate evaluations produced within both task-level resource limits. A task contributes to correctness when at least one candidate passes the numerical and CUDA execution checks. Its performance score is the largest retained speedup among these correct candidates. The active-time and cumulative output-token counts recorded at each evaluation determine eligibility. For interrupted responses with missing token usage, the response token limit provides an upper bound before budget eligibility is determined. Every scheduled task contributes to the denominator for correctness and fast$_p$.

All 178 tasks solved by the fine-tuned model and all four solved by the original checkpoint have retained timing measurements.

The one-epoch comparison retains three additional passing candidates whose recorded usage is below the limits but whose output-token upper bounds are unresolved because some responses lack usage metadata: two Trajectory SFT tasks on L2 and one Direct SFT task on L1. Their correctness checks and timing measurements are retained in the reported scores. Restricting scores to candidates with verified budget bounds gives Trajectory SFT correctness of 23\%, 6\%, and 0\% and fast$_1$ of 13\%, 3\%, and 0\% on L1--L3; Direct SFT scores are zero at all levels. All 31 Trajectory SFT and one Direct SFT successes have retained timing measurements.

\section{Published KernelBench comparisons}
\label{app:published-kernelbench}
Table~\ref{tab:main} places our learning study in the context of published KernelBench results. All rows retain the original authors' measurements. We summarize each source's generation language, hardware, inference budget, and candidate-selection procedure below.

The inference labels distinguish single-completion generation, a fixed multi-turn refinement workflow, and an agentic tool-use loop. CudaForge uses coding and judging agents within prescribed refinement rounds and is labeled multi-turn. PyTorch-reference results use the eager baseline unless a compiled mode is specified.

\paragraph{Direct generation.}
Qwen3-8B, DeepSeek-R1-0528, AutoTriton-8B, GPT-4o, and Claude Sonnet 4 values come from Table~2 of \citet{li2025autotriton}, using its Triton adaptation of KernelBench. We reproduce execution accuracy and fast$_1$. The paper does not specify the evaluation GPU or exact sampling budget. Kevin-32B and KernelCoder-32B values are the pass@1 results in Table~1 of \citet{kong2025concur}, evaluated on RTX~5090 GPUs against PyTorch eager. L3 is not reported there. Kevin was trained on 180 of the 200 L1/L2 tasks~\citep{baronio2026kevin}.

\paragraph{Iterative systems.}
CudaForge values are from Table~2 of \citet{zhang2025cudaforge}, covering the full benchmark on RTX~6000 GPUs with OpenAI-o3 as both coder and judge. The default budget is ten rounds, and the best correct candidate is selected. Claude Opus 4.5, Gemini 3 Pro, and CUDA Agent values are from Table~1 of \citet{dai2026cudaagent}, using the eager-reference faster-rate column. These models use the same agent loop, allowing up to 200 evaluation turns, a 131,072-token context, and H20 evaluation GPUs. CUDA Agent is trained from the proprietary Seed1.6 model. Candidate selection maximizes speedup over \texttt{torch.compile}; Table~\ref{tab:main} reports the eager-reference scores.

\paragraph{Additional performance metrics.}
Table~\ref{tab:threshold-results} combines AutoTriton's reported fast$_2$ values with our measured thresholds. ConCuR reports correctness and fast$_1$; CudaForge additionally reports speedup percentiles and geometric means; CUDA Agent reports geometric means and faster-than-compiled coverage.

\begin{table}[t]
\caption{KernelBench speedup coverage (\%), with $F_p$ denoting fast$_p$ relative to PyTorch eager. Published results use different evaluation protocols. The published rows generate Triton in a single turn; our rows generate CUDA with agentic inference.}
\label{tab:threshold-results}
\centering\small
\setlength{\tabcolsep}{7pt}
\renewcommand{\arraystretch}{1.1}
\begin{tabular}{@{}l rr rr rr@{}}
\toprule
Model & \multicolumn{2}{c}{L1} & \multicolumn{2}{c}{L2} & \multicolumn{2}{c}{L3} \\
\cmidrule(lr){2-3}\cmidrule(lr){4-5}\cmidrule(l){6-7}
 & $F_1$ & $F_2$ & $F_1$ & $F_2$ & $F_1$ & $F_2$ \\
\midrule
Qwen3-8B \citep{li2025autotriton} & 9 & 9 & 8 & 1 & 4 & 2 \\
DeepSeek-R1-0528 \citep{li2025autotriton} & 7 & 1 & 28 & 2 & 14 & 2 \\
AutoTriton-8B \citep{li2025autotriton} & 10 & 6 & 17 & 0 & 10 & 4 \\
GPT-4o \citep{li2025autotriton} & 3 & 1 & 3 & 0 & 4 & 2 \\
Claude Sonnet 4 \citep{li2025autotriton} & 11 & 7 & 10 & 1 & 2 & 0 \\
\midrule
\multicolumn{7}{@{}l}{\textit{Our Qwen3.5-9B comparison}} \\
Qwen3.5-9B & 0 & 0 & 0 & 0 & 0 & 0 \\
\quad + Direct SFT (1k) & 1 & 1 & 0 & 0 & 0 & 0 \\
\quad + Trajectory SFT (1k) & 13 & 6 & 4 & 2 & 0 & 0 \\
\quad + \kmdata{} SFT (full) & 64 & 31 & 62 & 51 & 12 & 6 \\
\bottomrule
\end{tabular}
\end{table}

\section{Dataset construction and composition}
\label{app:dataset-accounting}
\label{app:construction-details}
\subsection{Task inventory and provenance}
\begin{table}[t]
\caption{Composition of \kmdata{} (14,566 tasks). Each task has one development trajectory; train and validation contain disjoint task identifiers.}
\label{tab:dataset-inventory}
\centering\small
\begin{tabular}{lrrr}
\toprule
Input category & Dataset & Training & Validation \\
\midrule
Natural language & 354 & 334 & 20 \\
PyTorch code & 9,964 & 9,444 & 520 \\
Mixed language and code & 1,772 & 1,683 & 89 \\
Algebra/matrix & 451 & 430 & 21 \\
CUDA programming tasks & 2,025 & 1,924 & 101 \\
\midrule
Total & 14,566 & 13,815 & 751 \\
\bottomrule
\end{tabular}
\end{table}

Table~\ref{tab:dataset-inventory} partitions the 14,566 task--trajectory records from 41 sources by input category and split. Each task has one successful trajectory paired with an implementation that passes its evaluation checks; intermediate failed attempts remain part of the trajectory.

Task records preserve source attribution, interface requirements, and curation decisions. Generation records link candidate revisions, model configurations, selected solutions, and evaluation outcomes. Evaluation records distinguish reference-execution checks, restricted-buffer checks, and candidate tests. Some recovered controls transfer reference outputs to CUDA to exercise the candidate interface.

\subsection{Curation parameters}
\paragraph{Admission decisions.}
Individual reviews assess whether a task has a workable CUDA objective and a sufficiently specified computation. Known semantic defects and non-CUDA targets are excluded. The source-group admission policy requires at least 20 reviewed tasks, with more than 80\% scoring at least three on a zero-to-four workability scale. Tasks from an upstream validated collection also require verified extraction of their input specifications. Additional references from established libraries reuse source audits meeting the same workability-rate threshold and require an explicit CUDA implementation task. Individual defect findings override source-level admission evidence.

\paragraph{Selection parameters.}
Global diversity selection uses a cosine-similarity cutoff of $0.925$. The priority weight is the product of source, review, category, mathematical-reasoning, and CUDA-relevance factors. Source weights follow fixed provenance profiles; tested library references receive a factor of $1.08$. A recorded review score $q_i$ contributes $1+0.04\max(0,q_i-2)$, with a neutral factor of one for tasks without an individual score. Category factors are $1.10$, $1.00$, $1.08$, $1.12$, and $1.02$ for natural language, PyTorch code, mixed language/code, algebra/matrix, and CUDA programming, respectively. Mathematical tasks that leave a transformation for the model to discover receive a factor of $1.10$, and tasks labeled as primary kernel computations receive $1.02$; other tasks receive one for these factors. Ties are resolved by weight, prompt length, and task identifier.

Selection operates on one global embedding pool. A subsequent selection over mathematical tasks uses a cutoff of $0.97$, preserving the previously retained set and adding examples without a sufficiently similar retained representative. Category weights express selection preferences rather than fixed quotas.

\subsection{Test-suite construction}
We use GLM-5.3~\citep{glm53} to generate and repair executable test suites and implementation interfaces from the curated specifications and recovered upstream materials. Generated suites define test inputs, expected outputs, and numerical tolerances. Reference execution and candidate controls provide feedback for repairing test code and interfaces before trajectory generation.

\subsection{Trajectory generation}
We use Qwen3.8-Flash~\citep{qwen38flash} with reasoning enabled. The fixed generation policy allocates 100,000 output tokens and 3,600 active seconds for an initial attempt. Eligible continuations use a cumulative ceiling of 300,000 output tokens and up to 7,200 active seconds, retaining the earlier resource consumption. Each response is capped at 32,768 output tokens, and 300 seconds are reserved for finalization.

Evaluation is serialized per GPU to reduce interference, while generation and compilation can proceed concurrently. Compilation results are reused only for matching source and toolchain configurations. Stopping conditions and resource limits are applied at the trajectory level. Failures caused by the execution environment are distinguished from numerical mismatches.

\paragraph{Optional solution explanations.}
The framework can produce a standalone explanation from the public development history and selected source. This option requires the implementation to pass the public and held-out checks specified for that output format. Hidden outcomes are excluded from explanation generation. The resulting explanation summarizes the selected implementation and its development history. The trajectory-based training representation in this paper is described in Appendix~\ref{app:training-data}.

\subsection{Benchmark exclusion and duplicate checks}
To avoid benchmark contamination, we exclude benchmark splits and benchmark-associated repository paths during collection. Comparing the final dataset against all 250 KernelBench tasks yields no duplicates, including after removing comments, docstrings, and formatting differences.

\section{Dataset analysis}
\label{app:dataset-analysis}
We characterize computational vocabulary, task coverage, development context, supervision, and sensitivity to semantic selection. These analyses distinguish the diversity of specifications from the distribution of training examples. Related resources likewise examine domain composition, program length, and reasoning length~\citep{heakl2026cass,kong2025concur}; here, the unit of analysis additionally includes the development trajectory. Input categories follow the reviewed task labels. Trajectory and window analyses pool the training and validation splits.

\subsection{Computational vocabulary and co-occurrence}
\begin{figure}[!htbp]
\centering
\includegraphics[width=\linewidth]{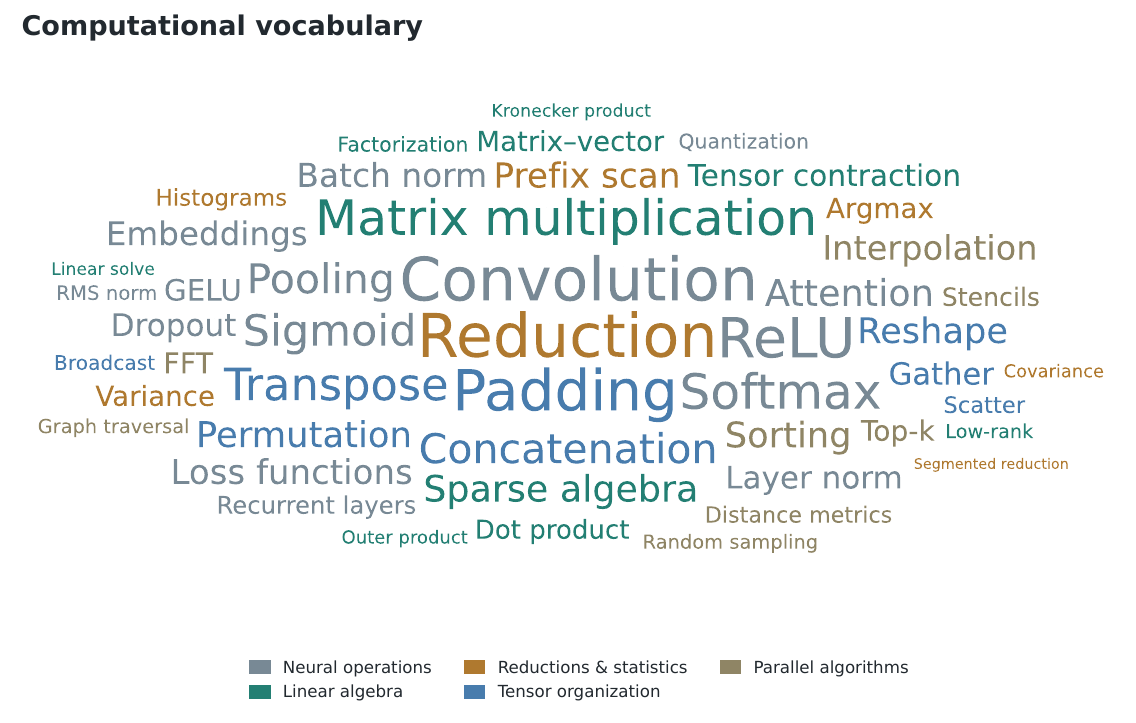}
\caption{Computational vocabulary across the 14,566 dataset task specifications. Word size increases with the square root of the number of tasks mentioning each descriptor; colors identify five thematic groups. Terms include both natural-language expressions and corresponding API names.}
\label{fig:dataset-vocabulary}
\end{figure}

\begin{figure}[!htbp]
\centering
\includegraphics[width=\linewidth]{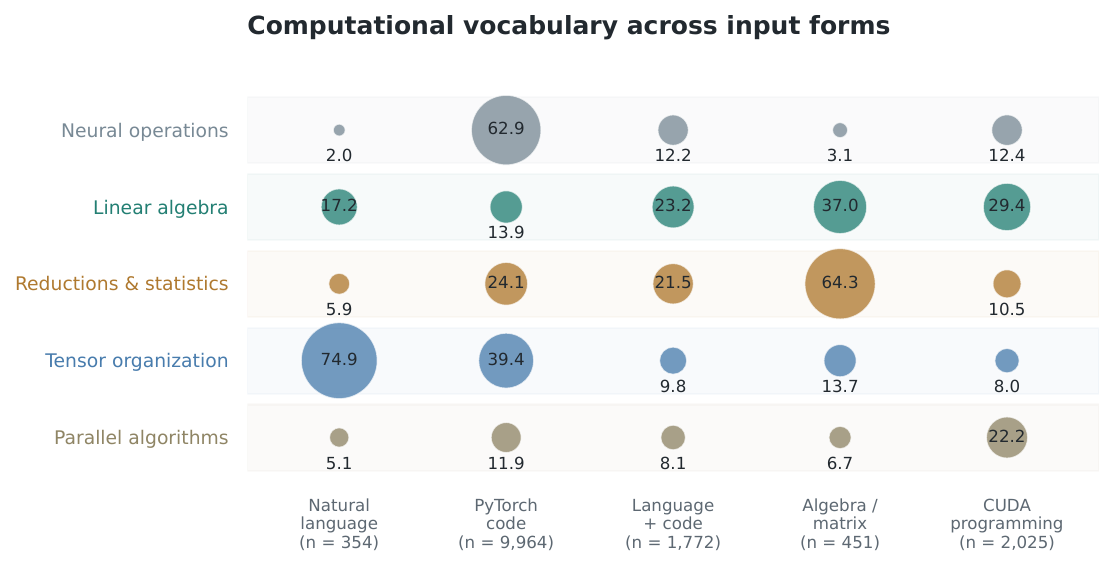}
\caption{Vocabulary across input forms. Circle area and numerical labels encode the percentage of tasks in each input category that mention at least one descriptor from the corresponding group. Groups may share tasks; each task is counted once per group. Column counts include all dataset tasks, including those without a matched descriptor.}
\label{fig:dataset-diversity-map}
\end{figure}

The specifications span neural operations, linear algebra, reductions, tensor organization, and parallel algorithms. Figure~\ref{fig:dataset-vocabulary} provides an overview through 48 computational descriptors, while Figure~\ref{fig:dataset-diversity-map} shows how these groups appear across input forms. Descriptors are matched against the visible task specification using a fixed vocabulary of terms and API names. Each task contributes at most once per descriptor or group, and a task may match several groups. Counts measure explicit mentions in the specification, including comments and constraints. Normalizing within each input form makes the smaller categories visible alongside the larger PyTorch collection.

\begin{figure}[!htbp]
\centering
\includegraphics[width=\linewidth]{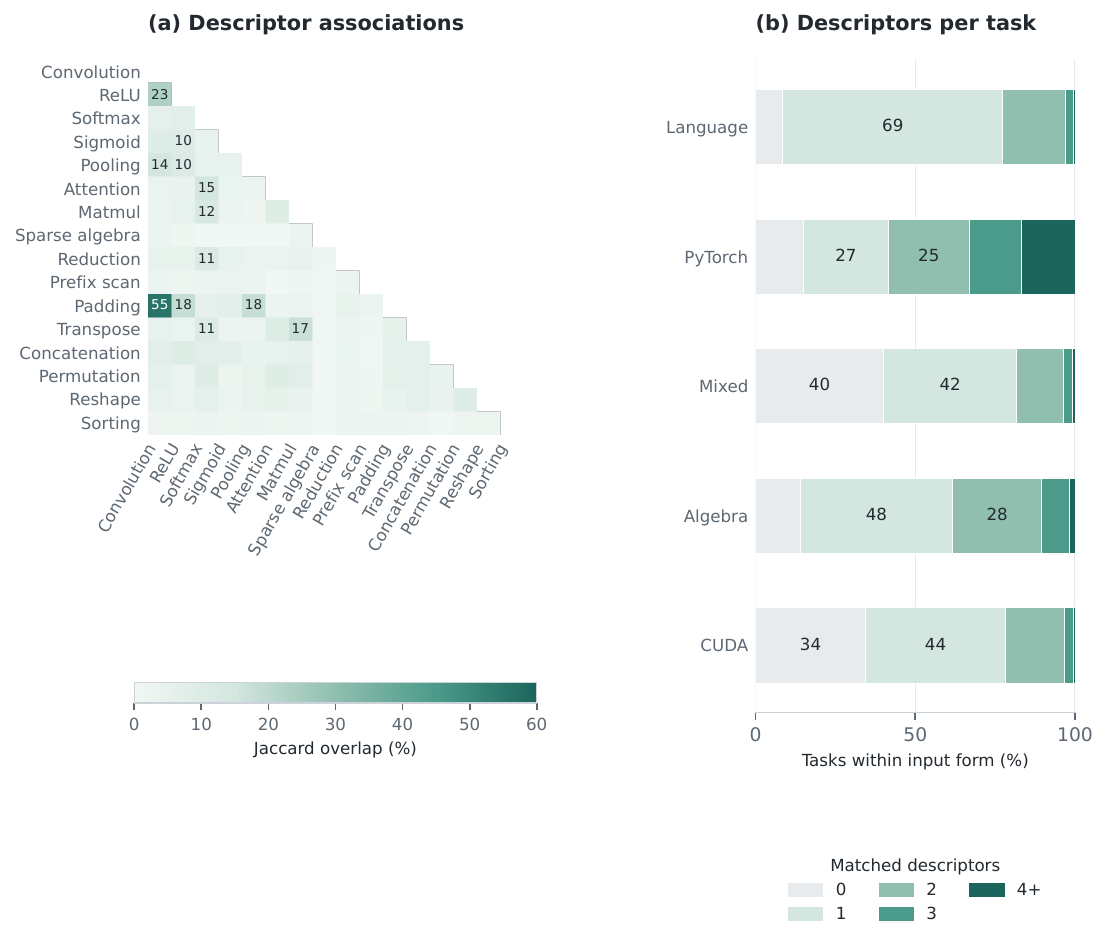}
\caption{Computational concepts within task specifications. (a) Jaccard overlap among the 16 most frequent descriptors, ordered by thematic group and frequency. Values of at least 10\% are labeled; the diagonal is omitted. (b) Number of distinct matched descriptors per task within each input category. The fixed lexical rules are applied to the task text, including code and comments; a zero denotes no match to the descriptor vocabulary. All 14,566 tasks are included.}
\label{fig:descriptor-composition}
\end{figure}

The vocabulary analysis also reveals combinations within individual tasks. Using the same fixed descriptor rules as Figure~\ref{fig:dataset-vocabulary}, 6,834 tasks (46.9\%) match two or more descriptors, and 5,736 (39.4\%) span at least two thematic groups. Multiple descriptors occur in 58.4\% of PyTorch tasks and 38.4\% of algebra tasks. Figure~\ref{fig:descriptor-composition} complements marginal descriptor counts with their co-occurrence structure.

We measure pairwise association by Jaccard overlap: the number of tasks matching both descriptors divided by the number matching either. Convolution and padding co-occur in 1,614 tasks, giving 55.3\% overlap. Matrix multiplication and transpose co-occur in 364 tasks (16.9\%), while attention and softmax co-occur in 264 (15.1\%). These associations characterize combinations of concepts expressed in the task specifications.

\subsection{Input coverage and source profiles}
\begin{figure}[!htbp]
\centering
\includegraphics[width=\linewidth]{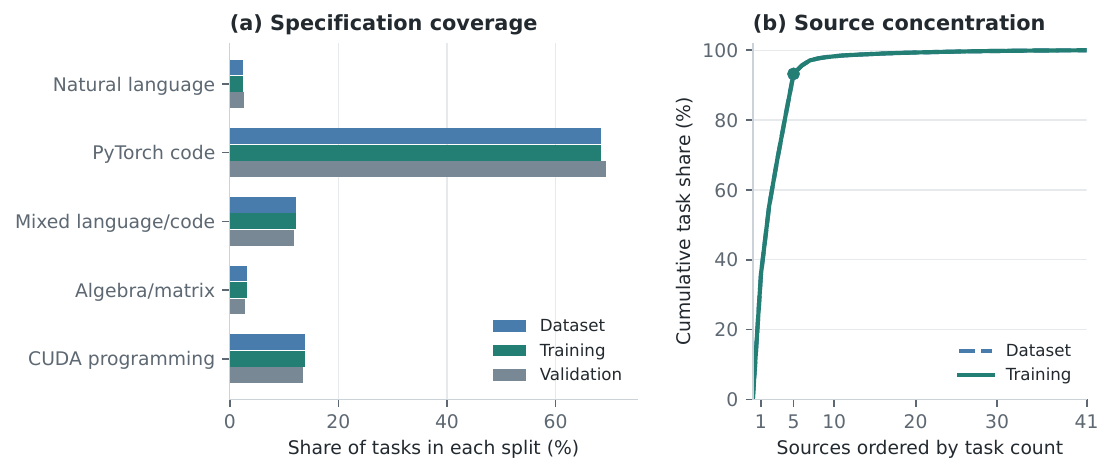}
\caption{Composition of the 14,566-task dataset. (a) Category proportions in the complete dataset and its training and validation splits. (b) Cumulative task share after independently ranking sources by size in the dataset and training split; markers identify the five largest sources.}
\label{fig:dataset-coverage}
\end{figure}

Figure~\ref{fig:dataset-coverage} compares input-category proportions and source concentration across the complete dataset and its splits. The training and validation splits have similar category proportions; Table~\ref{tab:dataset-inventory} gives the corresponding counts.

\begin{figure}[!htbp]
\centering
\includegraphics[width=\linewidth]{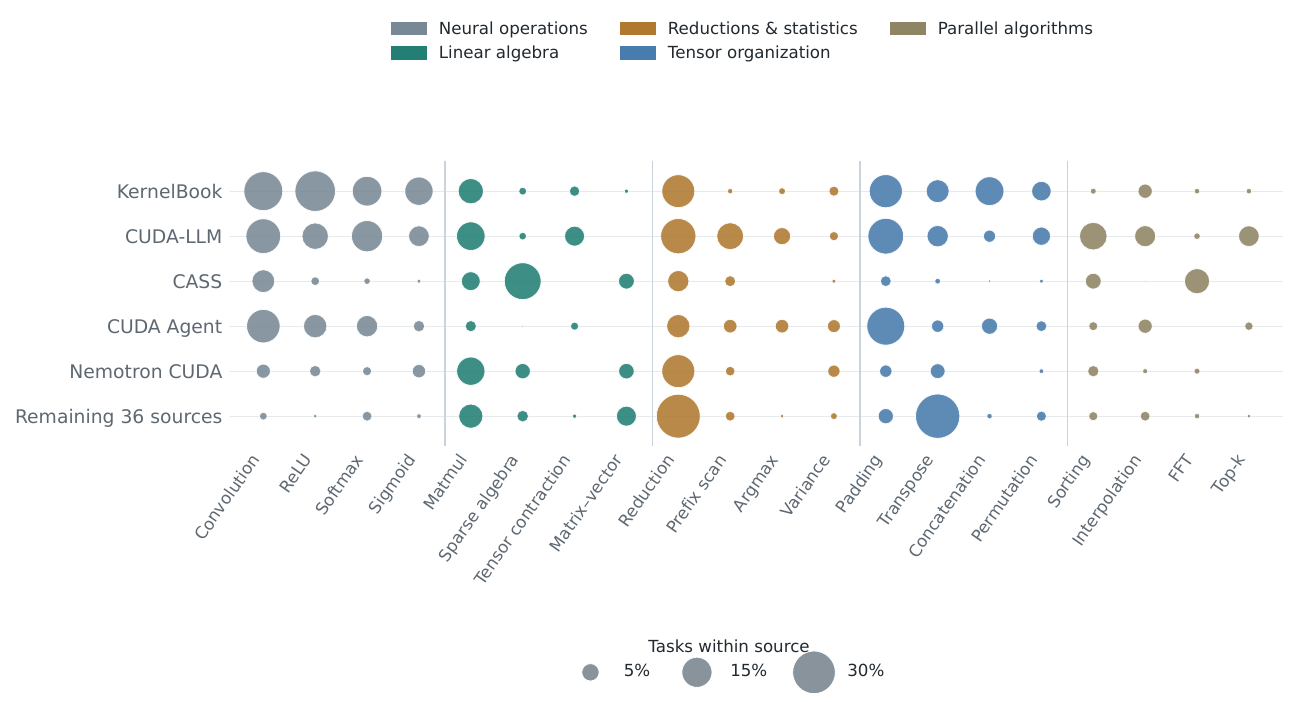}
\caption{Computational profiles of the collected sources. Circle area represents the percentage of tasks within a source matching a descriptor; colors indicate thematic groups. We display the four most frequent descriptors in each group, selected using the complete dataset. Source rows contain 5,219, 2,839, 1,935, 1,817, 1,772, and 984 tasks, respectively. The final row pools the remaining 36 sources. Source names refer to their contributions retained in \kmdata{}.}
\label{fig:source-profiles}
\end{figure}

Following source-composition analyses in SWE-smith~\citep{yang2025swesmith}, we also examine computational vocabulary within each source. Figure~\ref{fig:source-profiles} separates the five largest sources from the remaining 36, allowing their content profiles to be compared.

The profiles expose distinct concentrations. ReLU and convolution appear in 27.1\% and 24.9\% of KernelBook tasks, whereas sparse algebra and FFT appear in 22.3\% and 10.4\% of CASS tasks. Across the remaining sources, transpose and reduction appear in 32.3\% and 31.7\%. These differences give a concrete view of the complementary content brought together by collection across sources.

\subsection{Trajectory length and development depth}
\begin{figure}[!htbp]
\centering
\includegraphics[width=\linewidth]{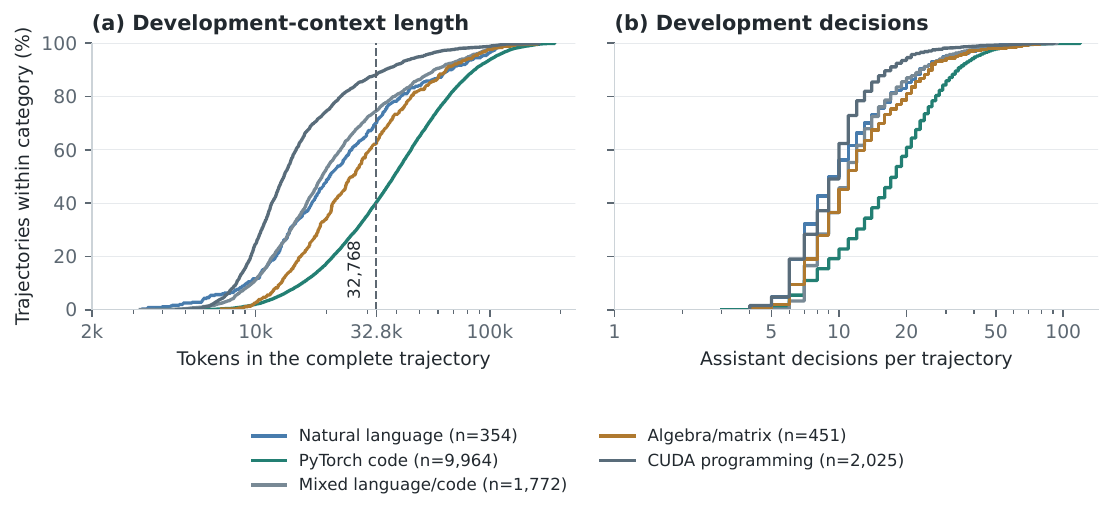}
\caption{Empirical trajectory distributions by input category, pooling both data splits. (a) Complete tokenized trajectory length before removal of historical reasoning. The dashed line marks the training context limit. (b) Number of assistant decisions, each comprising an assistant message and any associated tool observations. Curves normalize within each category; horizontal axes are logarithmic.}
\label{fig:dataset-trajectories}
\end{figure}

Figure~\ref{fig:dataset-trajectories} reports empirical cumulative distributions over all 14,566 trajectories. The median complete trajectory contains 30,904 tokens and 14 assistant decisions; the 90th percentiles are 79,308 tokens and 31 decisions. In total, 47.4\% exceed the 32,768-token training context limit. This distribution motivates segmenting long development trajectories into multiple training windows.

Trajectory length differs by specification category. PyTorch tasks have a median of 38,879 tokens and 18 decisions, compared with 13,203 tokens and 10 decisions for CUDA programming tasks. The difference also affects the number of windows derived from each task. Length measures the recorded development process, including prompts, reasoning, actions, and observations.

Figure~\ref{fig:development-depth} in Section~\ref{sec:dataset} relates trajectory length to interaction depth. In the shorter 16,384--32,768-token interval, median decision counts range from nine for natural-language tasks to twelve for PyTorch tasks, complementing the longer-trajectory comparisons in the main text.

\subsection{Supervision and context preservation}
\begin{figure}[!htbp]
\centering
\includegraphics[width=\linewidth]{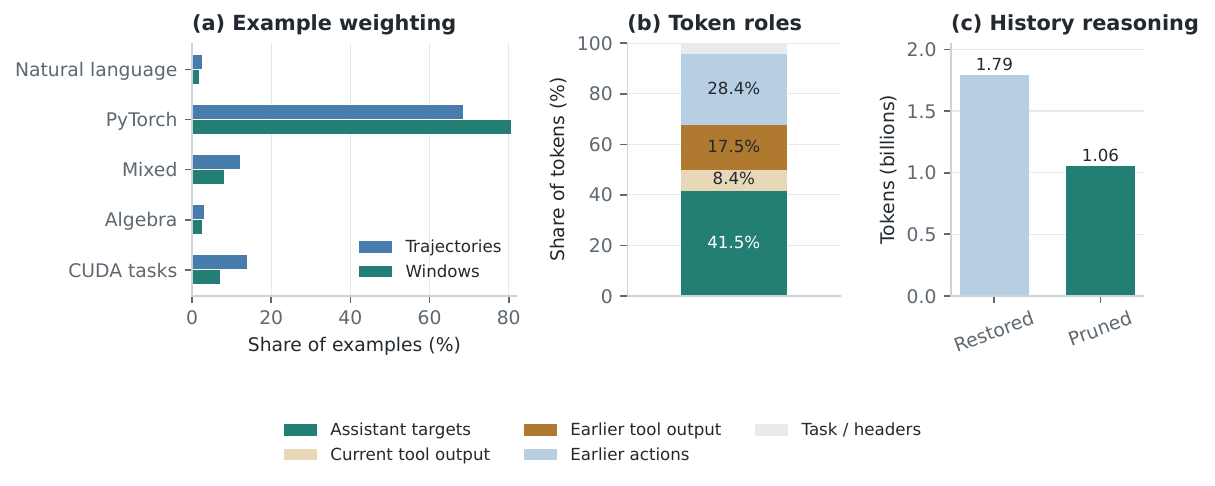}
\caption{How trajectories become learning examples. (a) Category shares among trajectories and derived windows. (b) Token roles across 1.057 billion tokens in both splits; only assistant targets contribute to the loss. Tool categories include message wrappers; assistant categories include message bodies and end tokens. (c) Token totals with historical reasoning restored or removed, holding windows and history truncation fixed. Token totals are measured per dataset copy.}
\label{fig:dataset-supervision}
\end{figure}

PyTorch tasks contribute 68.4\% of trajectories but 80.6\% of windows (Figure~\ref{fig:dataset-supervision}a). They produce 3.17 windows per trajectory on average, compared with 1.37 for CUDA programming tasks. Windowing therefore increases the share of examples from categories with longer trajectories.

Assistant targets comprise 41.5\% of all window tokens. The remaining tokens provide task and interaction context, including 26.0\% tool observations and 28.4\% earlier assistant actions. Removing historical reasoning reduces total tokens by 41.0\% relative to restoring that reasoning in the same windows with the same history truncation decisions (Figure~\ref{fig:dataset-supervision}c). Newly supervised reasoning remains intact. Appendix~\ref{app:training-data} specifies the construction and loss mask.

\subsection{Sensitivity of semantic selection}
\begin{figure}[!htbp]
\centering
\includegraphics[width=\linewidth]{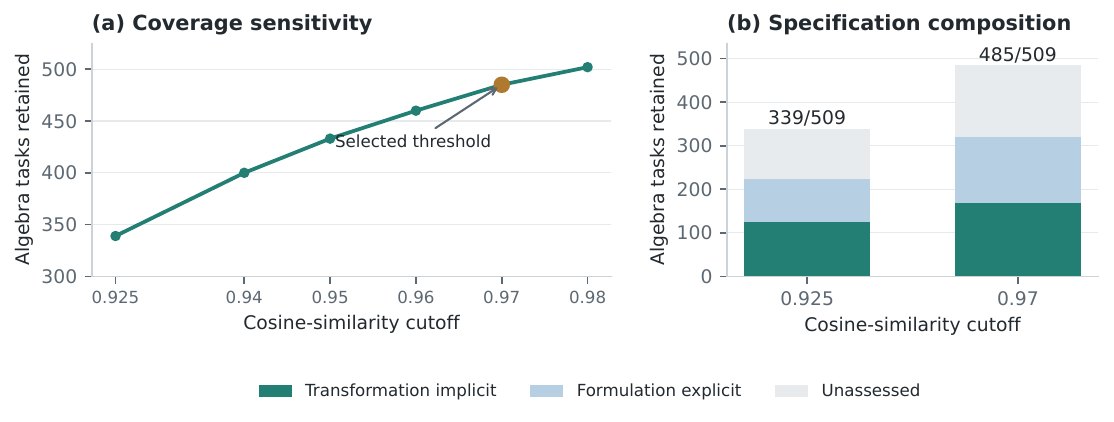}
\caption{Sensitivity of algebra coverage to the similarity cutoff, with the previously selected collection held fixed. (a) Six measured selection settings. (b) Retained specifications at the default and category-specific cutoffs, grouped by existing review annotations. Unassessed tasks lack this annotation. Counts refer to the pre-generation algebra candidate pool.}
\label{fig:dataset-curation}
\end{figure}

We examine the category-specific similarity threshold using a recorded algebra-selection sweep over a separate candidate pool before trajectory generation. Holding the previously selected collection fixed, we vary the cosine-similarity cutoff for admitting additional algebra tasks. A higher cutoff retains more candidates by requiring stronger similarity before treating a task as covered. Among 509 candidates, the retained set increases from 339 at 0.925 to 485 at the selected cutoff of 0.97 (Figure~\ref{fig:dataset-curation}). The added tasks include both explicit formulations and specifications that leave a useful transformation implicit.

\clearpage
\section{Model behavior on KernelBench}
\label{app:kernelbench-behavior}
We analyze all 500 evaluation trajectories: one from each checkpoint for every L1, L2, and L3 task. The analysis links archived tool actions with the candidate evaluations and resource counters used to compute the benchmark results.

\subsection{Evaluation failures and correctness repair}
The original Qwen3.5-9B checkpoint submits candidates for evaluation on 249 of 250 tasks, spanning 99 L1, 100 L2, and all 50 L3 tasks. Across its trajectories, 25,509 file writes complete successfully; calls to tool names outside the provided interfaces account for 97 of 35,323 actions (0.27\%).

At the first evaluation, 73 tasks fail because the candidate omits the required \texttt{cpp\_sources} argument to PyTorch's CUDA-extension loader, and 61 encounter module-state loading errors, including incompatible parameter names and incorrect \texttt{load\_state\_dict} signatures. Another 20 invoke PyTorch tensor computations in place of the required CUDA implementations. Other observed failures include undefined kernel entry points, compilation errors, and incorrect output shapes.

\begin{figure}[!htbp]
\centering
\includegraphics[width=\linewidth]{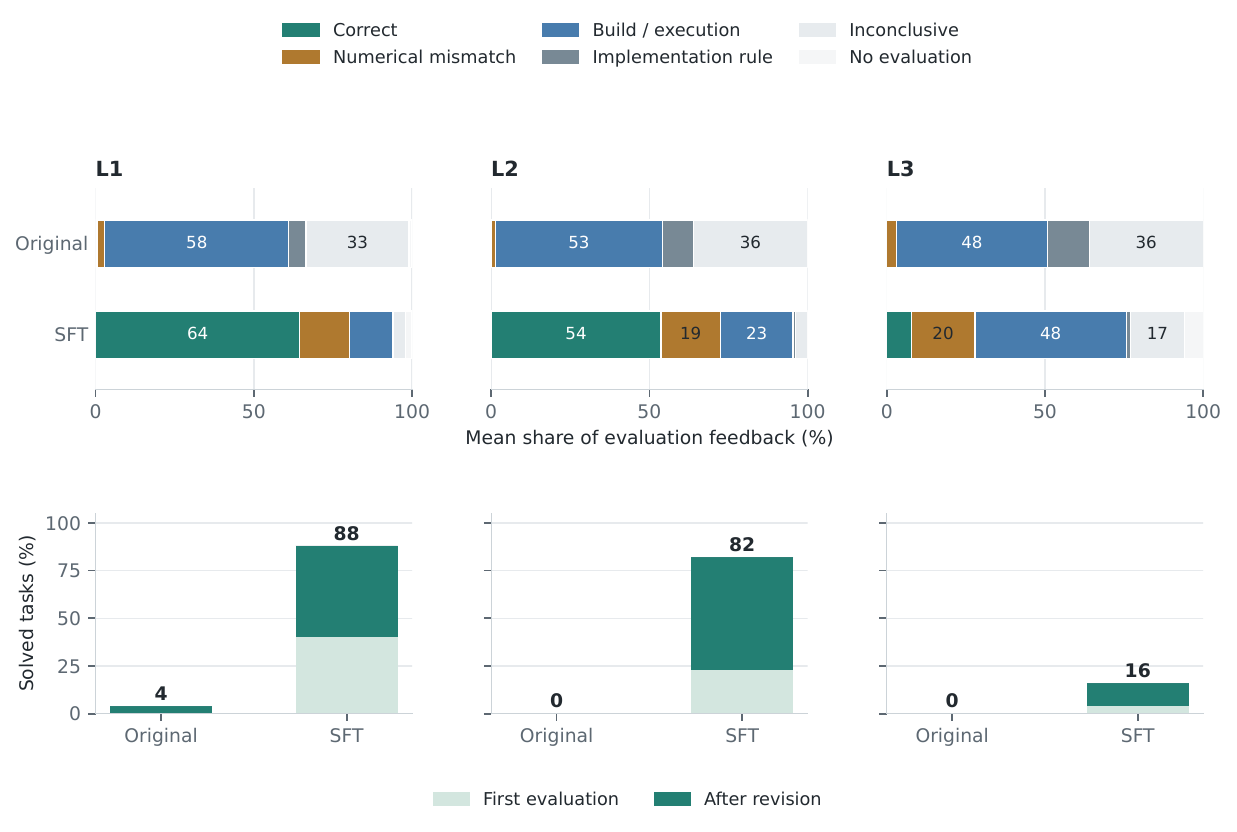}
\caption{Feedback and recovery under the common evaluation budget. Top: feedback proportions, normalized within each task and then averaged. Output mismatches include shape, value, and finiteness checks; build/execution errors include candidate construction, loading, and runtime failures; implementation-rule failures concern the CUDA execution requirement. Inconclusive checks remain a separate category. Tasks with no eligible evaluation are included explicitly. Bottom: correctness achieved at the first evaluation or after a revision, with all scheduled tasks in the denominator. Evaluations include recorded cache responses and repeated checks.}
\label{fig:feedback-recovery}
\end{figure}

Across complete trajectories, we classify feedback about outputs, execution, and CUDA implementation requirements, extending KernelBench's error analysis~\citep{ouyang2025kernelbench}. Figure~\ref{fig:feedback-recovery} summarizes each task's evaluation responses within the scoring budget. After SFT, correct evaluations account for 64.4\% and 53.6\% of the mean response distribution on L1 and L2, compared with 0.4\% and 0\% for the original model. Build or execution errors decrease from 58.2\% to 13.7\% on L1 and from 52.7\% to 23.0\% on L2. On L3, their shares remain similar (47.7\% and 47.8\%), while output mismatches account for 20.2\% after SFT. Thus, the change in feedback is strongest on individual operators and fused computations.

The first recorded evaluation passes on 40 L1, 23 L2, and two L3 tasks after SFT. Subsequent revisions solve a further 48, 59, and six tasks, respectively. In every such recovery, the first correct candidate differs from the initially evaluated candidate. Overall, 113 of 178 solved tasks (63.5\%) reach correctness through revision; on L2, the proportion is 59 of 82 (72.0\%). The original model's four successes also require revision, all on L1.

\subsection{Tool use and action sequences}
Each distinct tool call in the message archive contributes one action, including calls that return an error or remain interrupted. Figure~\ref{fig:kernelbench-behavior} first normalizes tool counts within each task and then averages the resulting proportions across tasks. This gives each workload equal weight. The Other category comprises tool names outside the five provided interfaces.

A repeated write is a completed write whose file path and content exactly match an earlier completed write in the same trajectory. We compare source contents by hash and count repetitions after the first occurrence. Table~\ref{tab:kernelbench-behavior} reports the fraction of all completed writes satisfying this criterion. Across the three levels, the original checkpoint repeats 15,322 of 25,509 completed writes (60.1\%), compared with 21 of 795 (2.6\%) after SFT. The numbers of tasks containing at least one repeated write are 229 and 14, respectively.

\begin{table}[htbp]
\caption{Trajectory behavior by model and KernelBench level. Calls is the median number of tool calls per task. Repeated writes is the percentage of completed writes reproducing an earlier file path and content. Profiling is the percentage of tasks invoking the profiler.}
\label{tab:kernelbench-behavior}
\centering\small
\setlength{\tabcolsep}{6pt}
\renewcommand{\arraystretch}{1.12}
\begin{tabular}{@{}llrrr@{}}
\toprule
Model & Level & Calls & Repeated writes & Profiling \\
\midrule
Original & L1 & 178.5 & 66.5\% & 6\% \\
Original & L2 & 99 & 53.1\% & 4\% \\
Original & L3 & 75 & 46.7\% & 2\% \\
\midrule
\kmdata{} SFT & L1 & 56 & 2.7\% & 87\% \\
\kmdata{} SFT & L2 & 71.5 & 2.2\% & 85\% \\
\kmdata{} SFT & L3 & 74 & 3.8\% & 30\% \\
\bottomrule
\end{tabular}
\end{table}

Table~\ref{tab:kernelbench-behavior} shows fewer tool calls on L1 and L2 alongside reduced repetition and increased profiling.

Action sequences reveal how an agent uses its tools together. Following the trajectory analyses of SWE-agent and SWE-smith~\citep{yang2024sweagent,yang2025swesmith}, we examine transitions between consecutive tool calls and the length of uninterrupted runs of the same tool type. For each current action, we normalize outgoing transitions within each task and average over tasks containing that transition context.

\begin{figure}[!htbp]
\centering
\includegraphics[width=.96\linewidth]{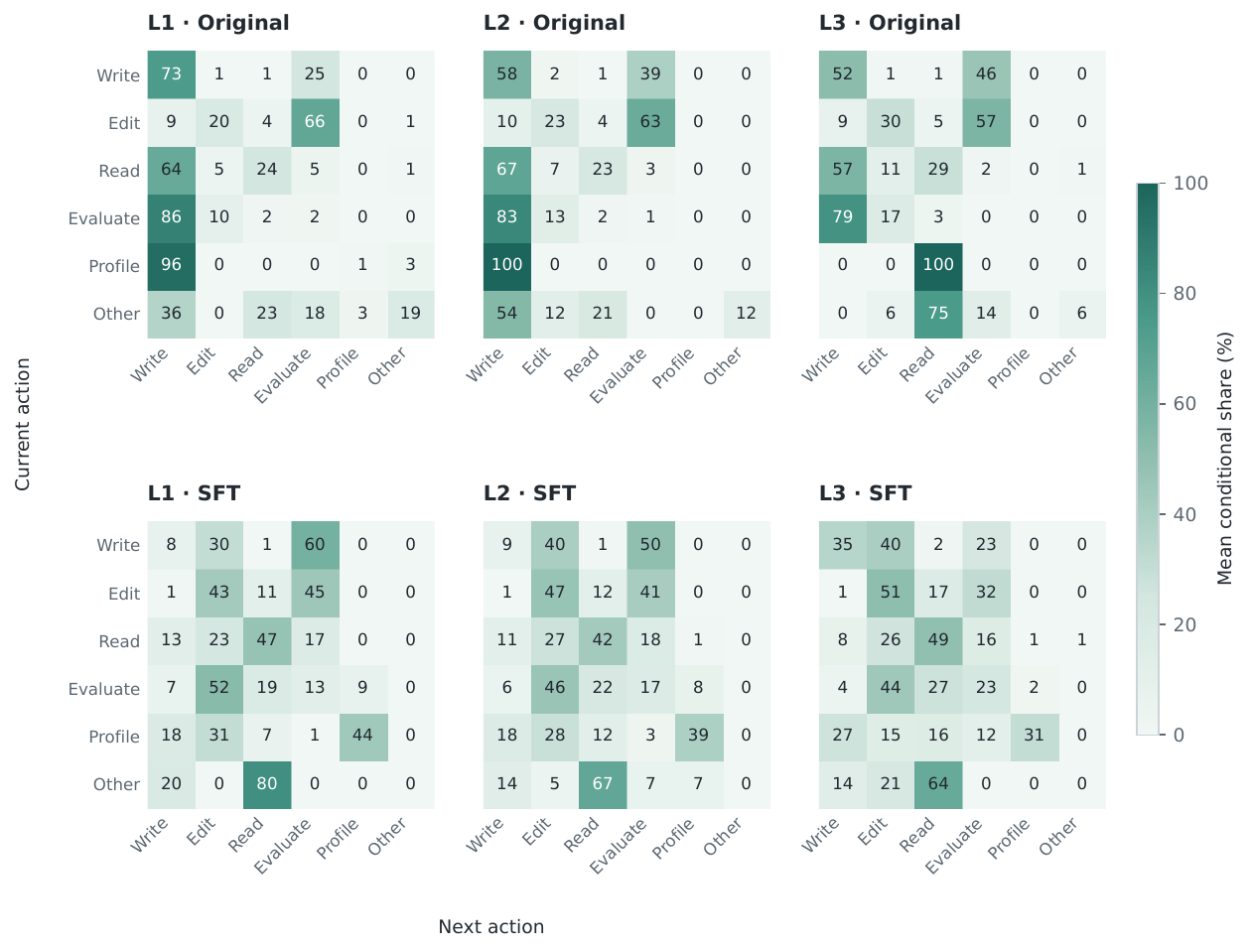}
\caption{Action transitions in original and SFT trajectories. Rows denote the current tool and columns the next tool; cells show percentages averaged over tasks after row normalization within each task. All archived tool calls are included, ordered by recorded start time. Each row averages only tasks with at least one outgoing transition from that tool. A dash denotes an absent context. Repetition here refers to tool type, whereas Table~\ref{tab:kernelbench-behavior} measures exact repeated write contents.}
\label{fig:action-transitions}
\end{figure}

Figure~\ref{fig:action-transitions} shows a consistent change across difficulty levels. After evaluation, the original model returns to writing a file in 86.2\%, 83.5\%, and 78.7\% of transitions on L1, L2, and L3. After SFT, these shares fall to 7.1\%, 6.2\%, and 4.2\%; editing becomes the most common next action, followed by code inspection on L1 and L2. The median longest run of a single tool type decreases from 92 to 5 calls on L1, from 16 to 5 on L2, and from 8.5 to 6 on L3. Across all levels, 167 original-model trajectories contain a run of at least ten calls to the same tool, compared with 16 SFT trajectories. These sequence-level measurements complement the aggregate tool proportions: SFT organizes development around shorter sequences of modifications, inspection, and evaluation.

\subsection{Candidate quality and subsequent optimization}
We define the first candidate as the implementation submitted at the first recorded evaluation; the agent may reason and edit before that submission. Table~\ref{tab:first-candidate} reports its correctness alongside the performance of the earliest passing implementation within the evaluation budget. These first correct implementations already outperform eager execution on 101 SFT tasks.

\begin{table}[htbp]
\caption{First-candidate correctness and first-correct-candidate performance on KernelBench. Correctness and fast$_1$ use all tasks at each level (100/100/50). Solved counts tasks with a passing candidate within budget. Median speedup is computed over each row's solved tasks, using their first passing implementation; a dash denotes no solved tasks.}
\label{tab:first-candidate}
\centering\small
\setlength{\tabcolsep}{6pt}
\renewcommand{\arraystretch}{1.12}
\begin{tabular}{@{}llrrrr@{}}
\toprule
& & First candidate & & \multicolumn{2}{c}{First correct candidate} \\
\cmidrule(lr){3-3}\cmidrule(l){5-6}
Model & Level & Correct (\%) & Solved & Median speedup & fast$_1$ (\%) \\
\midrule
Original & L1 & 0 & 4 & 0.19$\times$ & 0 \\
Original & L2 & 0 & 0 & --- & 0 \\
Original & L3 & 0 & 0 & --- & 0 \\
\midrule
\kmdata{} SFT & L1 & 40 & 88 & 1.04$\times$ & 49 \\
\kmdata{} SFT & L2 & 23 & 82 & 1.17$\times$ & 46 \\
\kmdata{} SFT & L3 & 4 & 8 & 1.17$\times$ & 12 \\
\bottomrule
\end{tabular}
\end{table}

\begin{figure}[!htbp]
\centering
\includegraphics[width=\linewidth]{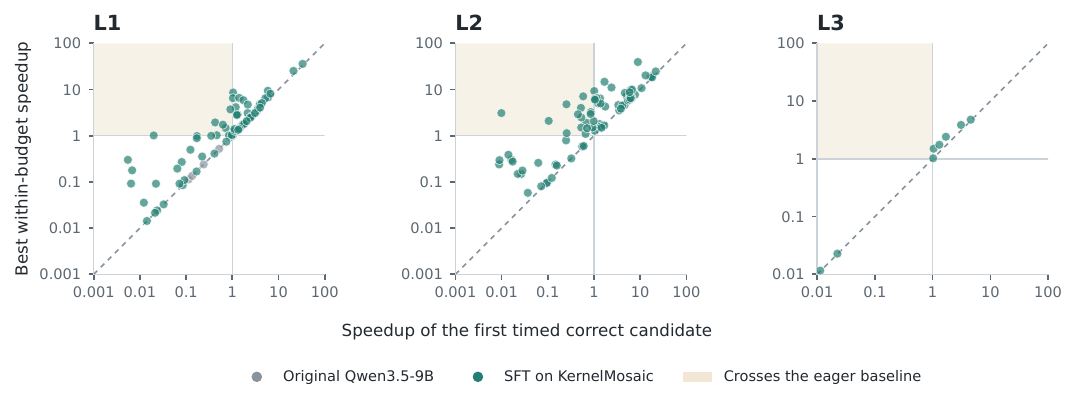}
\caption{Optimization after the first correct implementation. Each point compares a task's first timed correct candidate with its best measured candidate within the trajectory budget. Both axes show speedup over PyTorch eager on logarithmic scales; diagonal guides mark equal performance. The shaded region contains tasks that cross from slower-than-eager to faster-than-eager implementations. All 178 solved SFT tasks and four solved original-model tasks are included.}
\label{fig:kernelbench-optimization}
\end{figure}

For each solved task, Figure~\ref{fig:kernelbench-optimization} compares the first timed correct candidate with the best candidate found within budget, ordered by recorded active time. Further revisions bring an additional 15 L1 and 16 L2 tasks above the eager baseline. All six faster-than-eager L3 solutions already cross the baseline at their first timed correct candidate. The original model's four correct candidates remain slower than eager execution.

\subsection{Progress under token and time budgets}
\begin{figure}[!htbp]
\centering
\includegraphics[width=.96\linewidth]{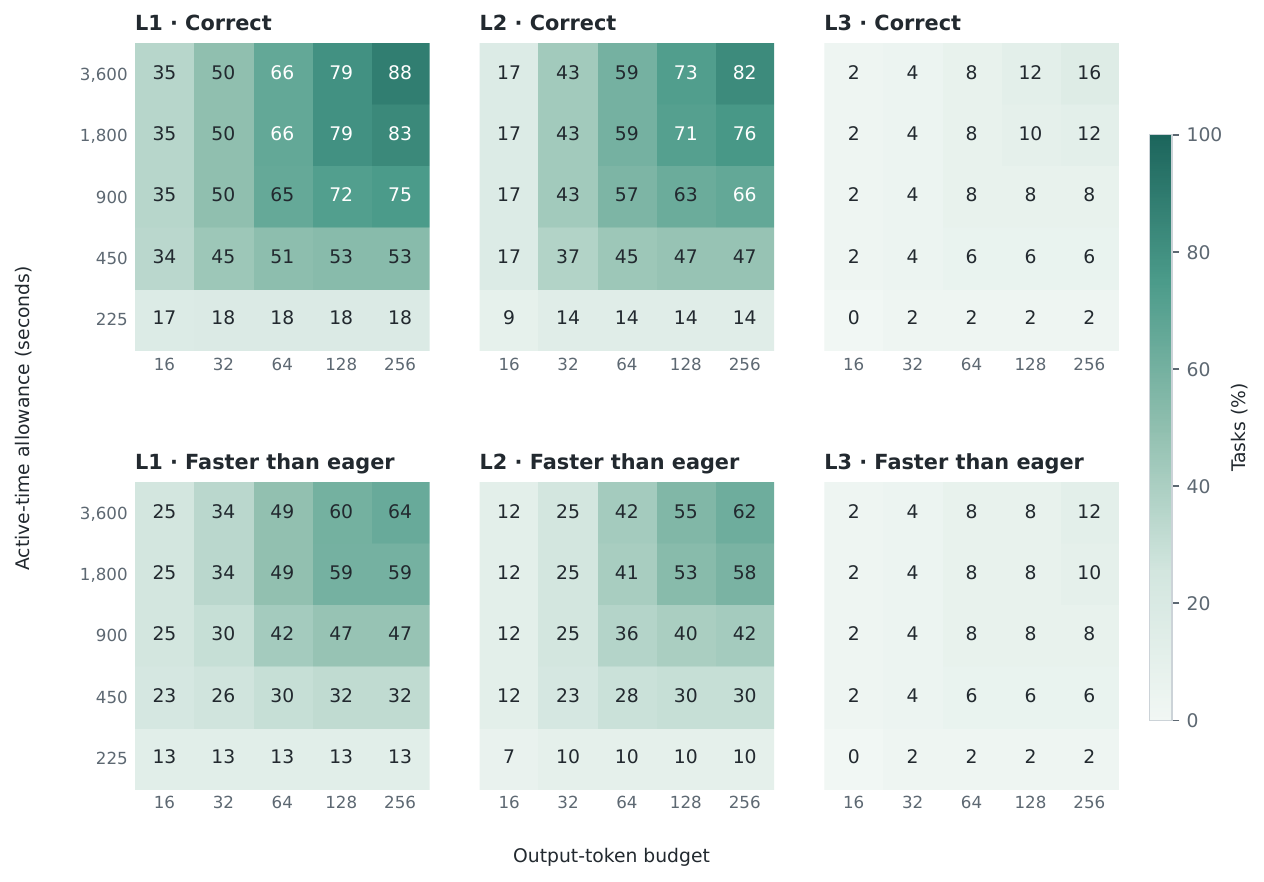}
\caption{SFT progress under joint resource cutoffs. Cells show the percentage of all tasks with a correct candidate (top) or a trusted faster-than-eager candidate (bottom) recorded before both cutoffs. Horizontal ticks denote 1,024 output tokens; active time excludes inference and infrastructure waits. All panels share the color scale.}
\label{fig:resource-surface}
\end{figure}

We apply output-token and active-time cutoffs to recorded candidate evaluations using the eligibility rules in Appendix~\ref{app:kernelbench-initial}. Correctness and fast$_1$ are computed over all scheduled tasks at each level. Figure~\ref{fig:kernelbench-behavior} varies the token cutoff with the time allowance fixed at 3,600 seconds; Figure~\ref{fig:resource-surface} varies both resources for the SFT trajectories.

At 65,536 output tokens, SFT solves 66\%, 59\%, and 8\% of L1, L2, and L3 tasks, with faster-than-eager coverage of 49\%, 42\%, and 8\%. At 131,072 tokens, correctness reaches 79\%, 73\%, and 12\%, and fast$_1$ reaches 60\%, 55\%, and 8\%. The original checkpoint solves one L1 task by 65,536 tokens and three by 131,072; its four solved tasks at the full budget are all in L1.

At 450 active seconds, L1 correctness reaches 53\% with 131,072 output tokens and remains unchanged at 262,144 tokens. Increasing the active-time allowance to 3,600 seconds at the latter token allowance raises correctness to 88\%. Faster-than-eager coverage continues to grow after many tasks have reached correctness: on L1, raising the active-time allowance from 900 to 3,600 seconds at the full token allowance increases correctness from 75\% to 88\% and fast$_1$ from 47\% to 64\%.

\subsection{Development activity after one epoch}
\label{app:sft-development}
\begin{figure}[!htbp]
\centering
\includegraphics[width=\linewidth]{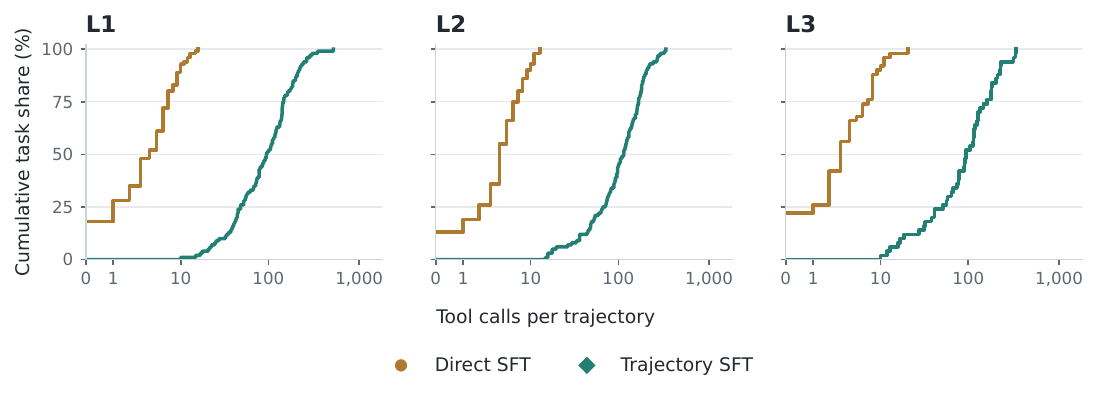}
\caption{Development activity after one epoch of SFT on 1,000 examples. Curves show cumulative task shares by the number of tool calls, including errored or interrupted invocations. Horizontal ticks show call counts on a $\log(1+\text{calls})$ scale. Each level includes all scheduled tasks.}
\label{fig:sft-development-depth}
\end{figure}

Across their 250 benchmark trajectories, Direct SFT and Trajectory SFT produce 1,133 and 29,253 archived tool calls, respectively (Figure~\ref{fig:sft-development-depth}). Median counts are 4, 4, and 3 for Direct SFT on L1--L3, versus 95.5, 114.5, and 94 for Trajectory SFT. Evaluation occurs on 55\%, 52\%, and 36\% of Direct SFT tasks, versus 98\%, 92\%, and 84\% after trajectory supervision; editing occurs on 18\%, 11\%, and 8\%, versus 98\%, 93\%, and 86\%. Trajectory supervision supports sustained development across all three levels.

\clearpage
\section{Illustrative dataset and development examples}
\label{app:examples}
These five examples illustrate the dataset's input forms. Descriptions are abridged and lightly paraphrased, preserving the computational target and material constraints while omitting boilerplate and supporting files. Table~\ref{tab:examples} records their origins.

\begin{table}[ht]
\caption{Illustrative prompts spanning the five curation categories. Input form is distinct from the requested implementation and from the evaluation interface.}
\label{tab:examples}
\centering\small
\begin{tabular}{p{0.23\linewidth}p{0.29\linewidth}>{\raggedright\arraybackslash}p{0.38\linewidth}}
\toprule
Input category & Example & Source / construction origin \\
\midrule
PyTorch code & Affine layer normalization & KernelBook~\citep{paliskara2025kernelbook} \\
Mixed language and code & Floating-point histogram & Nemotron SFT CUDA v1 \\
Natural language & Integer matrix transpose & fhrozen Stack CUDA prompts \\
Other CUDA code & Missing transform kernel & llama.cpp CUDA backend \\
Algebra / matrix & Batched low-rank operator & Synthetic algebra construction \\
\bottomrule
\end{tabular}
\end{table}

\subsection{A PyTorch reference: affine layer normalization}
The task requests an optimized CUDA replacement that preserves a PyTorch module's behavior and interface, returned as one Python module with CUDA bindings. Its forward computation is:
\begin{quote}\small
\begin{verbatim}
mean = x.mean(-1, keepdim=True)
s = (x - mean).pow(2).mean(-1, keepdim=True)
x = (x - mean) * torch.rsqrt(s + self.eps)
return self.scale * x + self.shift
\end{verbatim}
\end{quote}
The module initializes a learned scale to ones, a learned shift to zeros, and uses $\epsilon=10^{-5}$. The supplied input example has shape $(4,4,4,4)$ and four affine features. The computational specification comes primarily from the program; the wrapper states the target language and deliverable. Other shapes used in the development example below belong to its separately constructed adapter.

\subsection{Language and a header: a floating-point histogram}
This task from Nemotron-SFT-CUDA-v1~\citep{nvidia2026nemotroncuda} asks for configurable bins over a supplied minimum/maximum range, block-local accumulation in shared memory, and atomic accumulation into global memory, with support for arbitrary element and bin counts. The supplied header declares:
\begin{quote}\small
\begin{verbatim}
void compute_histogram(const float* input, int* histogram,
                       int num_elements, int num_bins,
                       float min_val, float max_val);
\end{verbatim}
\end{quote}
The requested deliverable is a CUDA source file using the supplied headers without modification. Here prose supplies requirements that the signature alone cannot express, while the header fixes the interface.

\subsection{A language specification: integer matrix transpose}
This prompt from the CUDA subset of Stack-Prompts~\citep{fhrozen2026stackprompts} specifies a kernel named \texttt{naiveGPUTranspose} with integer input/output pointers and \texttt{rows}, \texttt{cols} arguments. Each thread obtains a two-dimensional position $(i,j)$, checks $i<\texttt{rows}$ and $j<\texttt{cols}$, and writes
\begin{equation*}
\texttt{d\_b}[j\cdot\texttt{rows}+i]
=\texttt{d\_a}[i\cdot\texttt{cols}+j].
\end{equation*}
The original specification explicitly requests one element per thread and standard block/thread indexing. Thus the collection includes implementation-constrained instructions as well as open-ended optimization tasks.

\subsection{A source-completion task: a Walsh--Hadamard transform}
The source from llama.cpp's CUDA backend~\citep{llamacpp2026} leaves the definition of \texttt{fwht\_cuda<N>} missing while preserving its signature, launch constraints, and caller. The requested output is the complete missing definition. The context supplies contiguous tensors of equal shape, row counts, four rows per block, a scale factor $1/\sqrt{N}$, and specializations for $N\in\{64,128,256,512\}$. The kernel receives \texttt{src}, \texttt{dst}, \texttt{n\_rows}, and \texttt{scale}. Existing helper conventions constrain the reconstruction. This example represents integration with an existing CUDA codebase, where correctness depends on both the transformation and its surrounding call contract.

\subsection{A mathematical specification: a batched low-rank operator}
The task supplies contiguous, row-major float32 tensors $U,V\in\mathbb{R}^{G\times Q\times R}$ and $X\in\mathbb{R}^{G\times Q\times C}$. For each block $b$, it defines
\begin{equation*}
L_b=U_bV_b^\top,\qquad Y_b=L_bX_b,
\end{equation*}
and requests a float32 output $Y\in\mathbb{R}^{G\times Q\times C}$. All dimensions are positive, $C\geq2$, and inputs must remain unchanged. Accumulation uses float32 or higher precision, with elementwise absolute and relative tolerances of $10^{-4}$ against the float64 specification. The requested module exposes \texttt{solve(U, V, X, *, dimensions)} and uses custom CUDA kernels through PyTorch bindings. The prompt defines the operator without prescribing the multiplication order. This leaves implementation choices available while making types, shapes, and numerical obligations explicit.

\subsection{An observed implementation revision}
\label{app:layernorm}
A separate five-task inspection run used Qwen3.8-Flash to develop implementations for these examples. The run used an example-specific adapter with 32 public and 25 hidden input cases across the five tasks. All five selected implementations passed that adapter's public and hidden checks.

\textbf{Initial implementation.} Round 0 assigns one block to each row, with a fixed 256-thread block. It computes the mean, then the centered sum of squares, then normalizes and applies the affine parameters. Two separate block reductions compute the mean and variance, with FP32 accumulation. This version passed all three public cases.

\textbf{Feedback and revision.} The next-round record uses public timing and source inspection to hypothesize that small rows incur avoidable reduction and synchronization overhead. Round 1 jointly accumulates the sum and sum of squares, reduces both quantities together, and computes
\begin{equation*}
\mu=\frac{1}{N}\sum_i x_i,\qquad
\widehat{\sigma}^{2}=\max\!\left(0,\frac{1}{N}\sum_i x_i^2-\mu^2\right).
\end{equation*}
It chooses a warp-multiple block size, capped at 256 threads, from the row length. The binding also adds explicit contiguity checks before materialization. The joint reduction uses two synchronization barriers. Table~\ref{tab:layernorm-case} gives the recorded outcomes, including hidden evaluation of the selected revision.

\begin{table}[ht]
\caption{LayerNorm implementation revision. Latencies are task-level aggregates from the adapter, in microseconds. Public rounds use the same public cases; the last row uses separate hidden cases.}
\label{tab:layernorm-case}
\centering\small
\begin{tabular}{llrrr}
\toprule
Candidate / evaluation & Correctness & Candidate ($\mu$s) & Eager ($\mu$s) & Speedup \\
\midrule
Round 0 / public & 3/3 cases & 16.77 & 72.28 & 4.31$\times$ \\
Round 1 / public & 3/3 cases & 10.76 & 48.94 & 4.55$\times$ \\
Selected round 1 / hidden & 5/5 cases & 11.83 & 48.95 & 4.14$\times$ \\
\bottomrule
\end{tabular}
\end{table}

\end{document}